\def\year{2022}\relax
\documentclass[letterpaper]{article} % DO NOT CHANGE THIS
\usepackage{aaai22}  % DO NOT CHANGE THIS

\usepackage{times}  % DO NOT CHANGE THIS
\usepackage{helvet}  % DO NOT CHANGE THIS
\usepackage{courier}  % DO NOT CHANGE THIS
\usepackage[hyphens]{url}  % DO NOT CHANGE THIS
\usepackage{graphicx} % DO NOT CHANGE THIS
\usepackage{natbib}  % DO NOT CHANGE THIS AND DO NOT ADD ANY OPTIONS TO IT
\usepackage{caption} % DO NOT CHANGE THIS AND DO NOT ADD ANY OPTIONS TO IT
\DeclareCaptionStyle{ruled}{labelfont=normalfont,labelsep=colon,strut=off} % DO NOT CHANGE THIS
\usepackage{algorithm}
\usepackage{algorithmic}
\usepackage{graphicx}
\usepackage{amsmath}
\usepackage{amssymb}
\usepackage{subfigure}
\usepackage{float}
\usepackage{color}
\usepackage{colortbl}
\usepackage{verbatim}
\usepackage{tabularx}
\usepackage{multirow}
\usepackage[switch]{lineno}  %

\makeatletter
\def\hlinewd#1{%
\noalign{\ifnum0=`}\fi\hrule \@height #1 \futurelet
\reserved@a\@xhline}

\usepackage{newfloat}
\usepackage{listings}
\floatstyle{ruled}
\newfloat{listing}{tb}{lst}{}
\floatname{listing}{Listing}
\title{Deep Translation Prior: Test-time Training for Photorealistic Style Transfer}
\author{
Sunwoo~Kim\thanks{Equal Contribution.} \quad Soohyun~Kim$^{*}$ \quad Seungryong Kim\thanks{Corresponding author.}\\
}

\affiliations{
Korea University, Seoul, Korea\\
\{sw-kim, shkim1211, seungryong\_kim\}@korea.ac.kr
}

\usepackage[utf8]{inputenc} % allow utf-8 input
\usepackage[T1]{fontenc}    % use 8-bit T1 fonts
\usepackage{hyperref}       % hyperlinks
\usepackage{url}            % simple URL typesetting
\usepackage{booktabs}       % professional-quality tables
\usepackage{amsfonts}       % blackboard math symbols
\usepackage{nicefrac}       % compact symbols for 1/2, etc.
\usepackage{microtype}      % microtypography
\usepackage{xcolor}         % colors
\usepackage{times}
\usepackage{epsfig}
\usepackage{graphicx}
\usepackage{amsmath}
\usepackage{amssymb}
\usepackage{subfigure}
\usepackage{float}
\usepackage{color}
\usepackage{colortbl}
\usepackage{verbatim}
\usepackage{tabularx}
\usepackage{multirow}
\usepackage{cuted}

\newcommand{\figref}[1]{Fig. \ref{#1}}
\newcommand{\tabref}[1]{Table \ref{#1}}

\definecolor{shcolor}{rgb}{0.5, 0.0, 1.0}

\begin{document}

\maketitle

\begin{abstract}
Recent techniques to solve photorealistic style transfer within deep convolutional neural networks (CNNs) generally require intensive training from large-scale datasets, thus having limited applicability and poor generalization ability to unseen images or styles. To overcome this, we propose a novel framework, dubbed Deep Translation Prior (DTP), to accomplish photorealistic style transfer through test-time training on given input image pair with untrained networks, which learns an image pair-specific translation prior and thus yields better performance and generalization. Tailored for such
test-time training for style transfer, we present novel network architectures, with two sub-modules of correspondence and generation modules, and loss functions consisting of contrastive content, style, and cycle consistency losses. Our framework does not require offline training phase for style transfer, which has been one of the main challenges in existing methods, but the networks are to be solely learned during test-time. Experimental results prove that our framework has a better generalization ability to unseen image pairs and even outperforms the state-of-the-art methods.
\end{abstract}

\section{Introduction}
Photorealistic style transfer is one of appealing image manipulation and editing tasks, which aims to, given a pair of images, i.e., the content and style image, synthesize an image by transferring the style to the content. Recent approaches for this task leverage statistics of content and style features extracted by deep convolutional neural network (CNNs)~\cite{gatys2016image,li2017universal,ulyanov2017improved}, which can be divided into \emph{optimization}-based and \emph{learning}-based methods. Optimization-based methods~\cite{gatys2016image,li2016combining,luan2017deep} directly obtain a stylized image by optimizing an image itself with well-defined content and style loss functions. As the seminal work, Gatys et al.~\cite{gatys2016image} present the style loss function based on Gram matrix and optimize the stylized image with the loss function, of which many variants were also proposed~\cite{li2016combining,gatys2016image,luan2017deep}. Since the loss function for style transfer is often non-convex, most methods leverage an iterative solver to optimize the output image itself~\cite{li2016combining,gatys2016image,luan2017deep}, and thus they can benefit from error feedback for stylization. Moreover, they are limited to encode an image translation prior on synthesized images, and thus often generate artifacts and show limited photorealism.

In contrast, recent learning-based methods~\cite{li2017universal,li2018closed,gu2018arbitrary,yoo2019photorealistic,huang2017arbitrary, park2019arbitrary} attempt to address these limitations by learning such image translation prior within networks from large-scale datasets~\cite{deng2009imagenet,lin2014microsoft}, often followed by pre- or post-processing~\cite{li2018closed, yoo2019photorealistic, huang2017arbitrary}.
Since it is notoriously challenging to collect training pairs for photorealistic style transfer due to its subjectivity, most methods alternatively leverage an auto-encoder to learn a decoder that captures the translation prior~\cite{huang2017arbitrary,chen2016fast,li2017universal}. However, during the training, these methods~\cite{li2018closed,li2019optimal} do not leverage explicit content and style loss functions, and thus may have poor generalization ability on unseen images or styles. In addition, adopting fixed network parameters at test-time may not account for the fact that a pair of images may require their own prior, namely an image pair-specific translation prior.
\begin{figure*}[t!]
	\centering
	\renewcommand{\thesubfigure}{}
	\subfigure[(a)]
	{\includegraphics[width=0.33\linewidth]{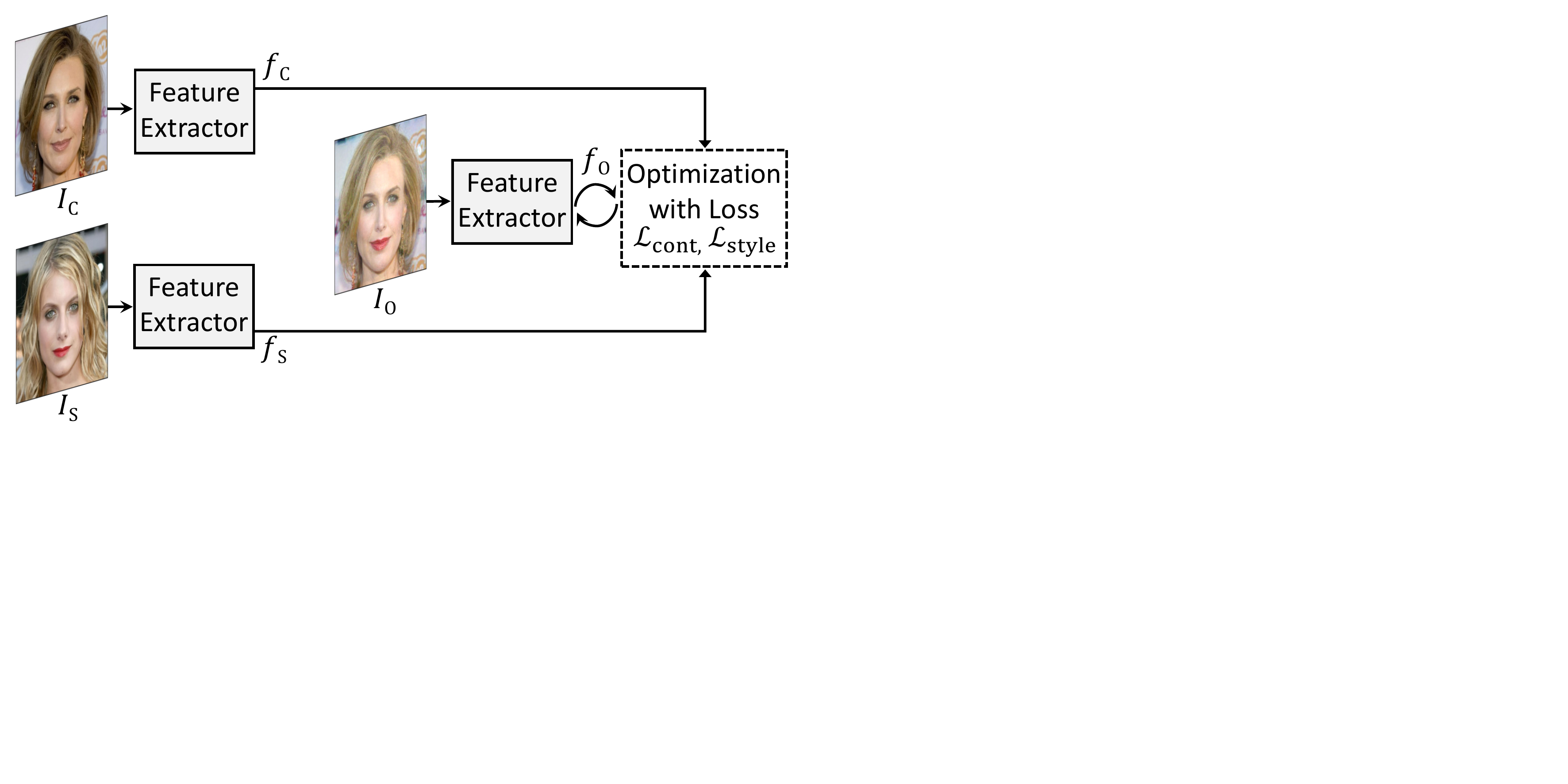}}\hfill
	\subfigure[(b)]
	{\includegraphics[width=0.33\linewidth]{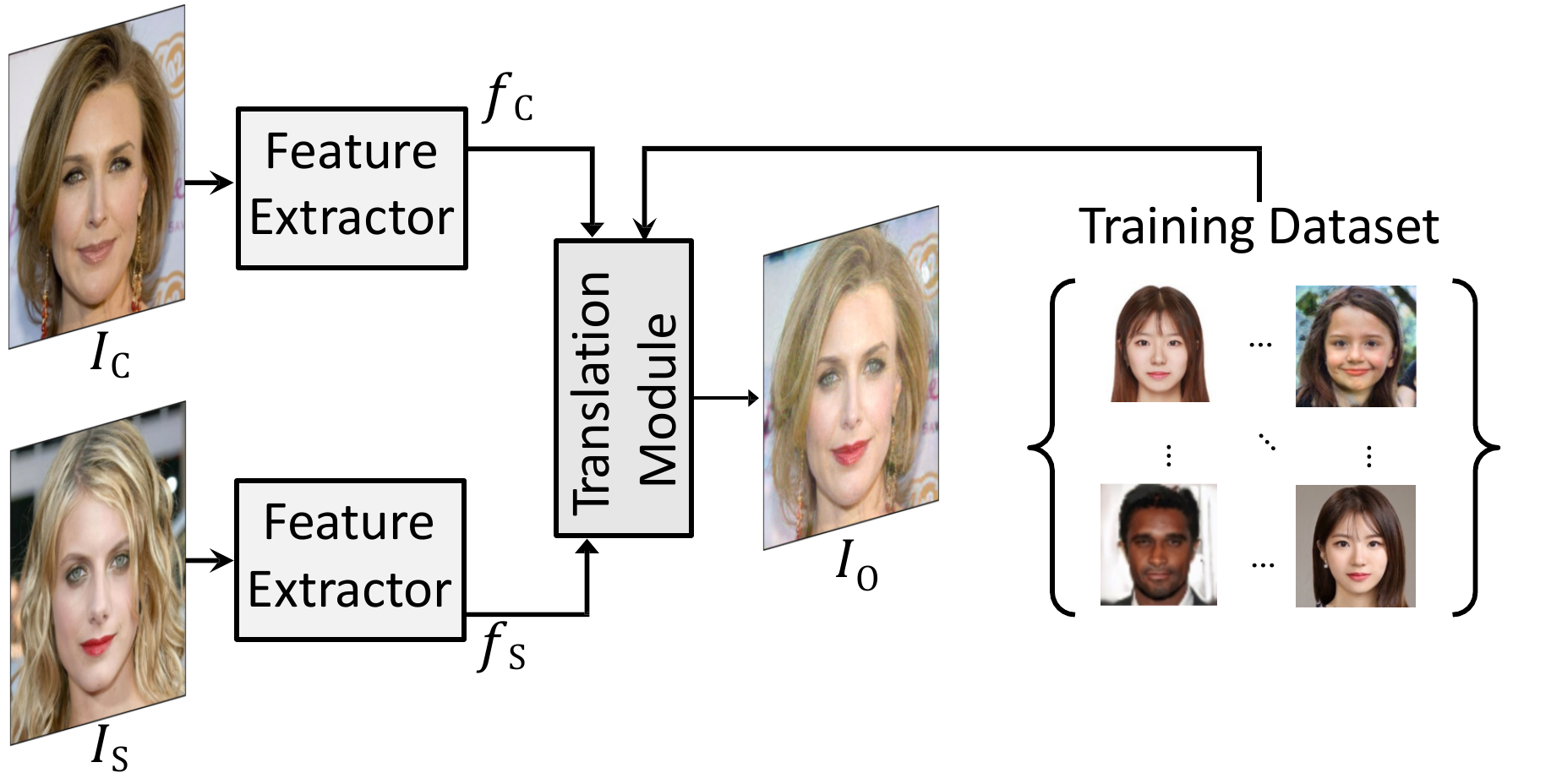}}\hfill
	\subfigure[(c)]
	{\includegraphics[width=0.33\linewidth]{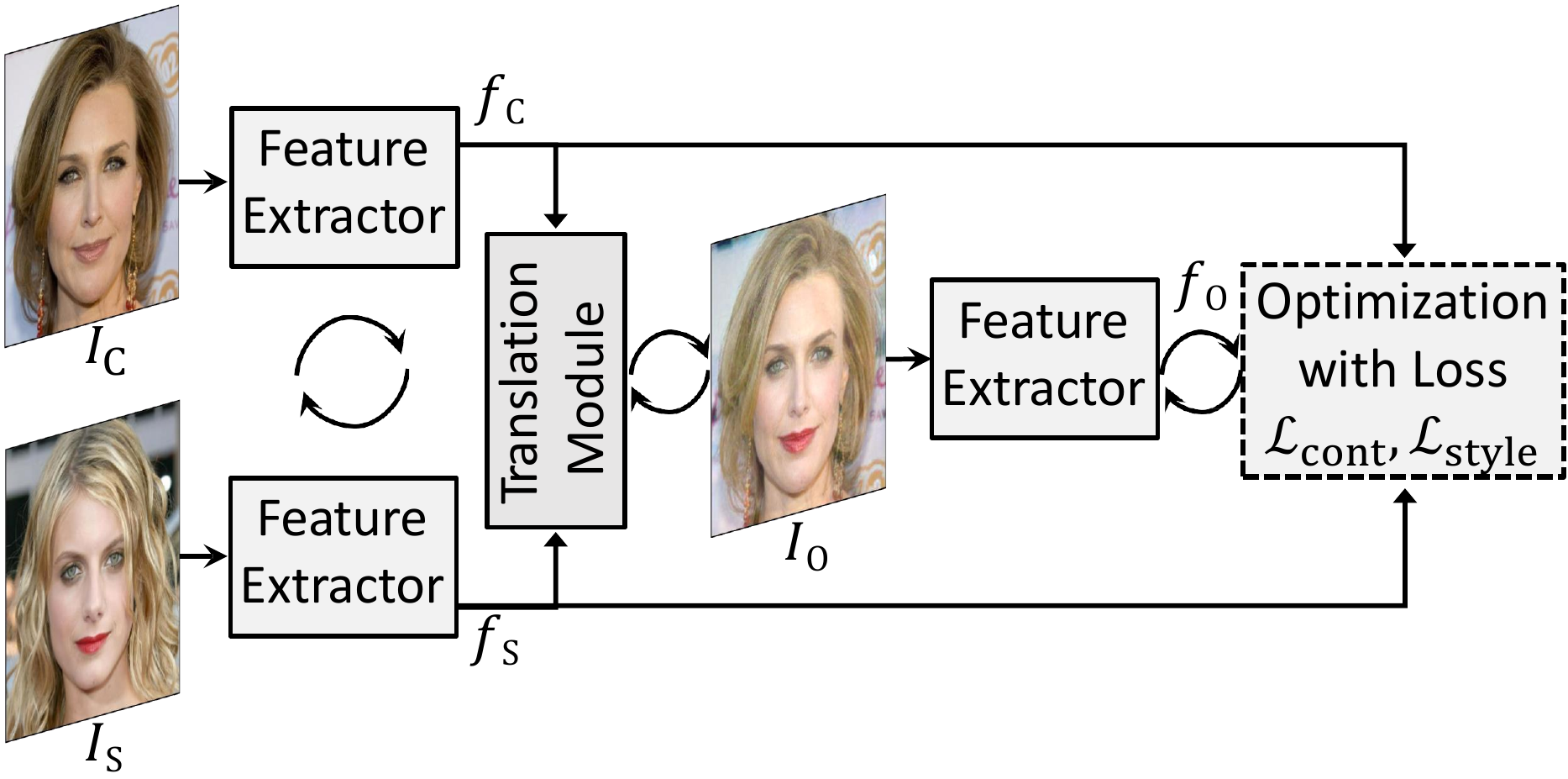}}\hfill\\
	\vspace{-10pt}  
	\caption{\textbf{Intuition of DTP:} (a) conventional optimization-based methods~\cite{li2016combining,gatys2016image,luan2017deep} that optimize an output \textbf{\emph{image itself at test-time}} with explicit content loss and style loss, which often generate artifacts and produce limited photorealism due to the lack of translation prior, (b) recent learning-based methods that require \textbf{\emph{intensive offline training}} from large-scale training data and use \textbf{\emph{pretrained and fixed}} networks at test-time~\cite{huang2017arbitrary,li2017universal,li2018closed,gu2018arbitrary,yoo2019photorealistic}, and (c) DTP that  \textbf{\emph{learns the untrained networks at test-time on an input image pair}} to capture an image pair-specific translation prior and thus provides better generalization to unseen images or styles.}\label{img:2}\vspace{-10pt}    
\end{figure*} 

In this paper, we explore an alternative, dubbed Deep Translation Prior (DTP), to overcome aforementioned limitations of both optimization- and learning-based methods. Our work accomplishes this without need of intensive training process using large-scale dataset or paired data, but through a test-time training on given input image pair. 
We argue that the translation prior does not necessarily need to be learned from intensive learning. Instead, an image pair-specific translation prior can be captured by solely minimizing explicit content and style loss functions on the image pair with untrained network for stylization. Tailored to this framework, we formulate novel network architectures consisting of two sub-modules, namely correspondence and generation modules, which are learned with well-designed content, style, and cycle consistency loss functions at test-time.

Our experiments on standard benchmark for photorealistic style transfer~\cite{luan2017deep,an2019ultrafast}, CelebA-HQ~\cite{liu2015deep}, and Flickr Faces HQ (FFHQ)~\cite{karras2019style} demonstrate that our framework consistently outperforms the existing methods.

\section{Related Work}
\paragraph{Style Transfer.}
Traditional optimization-based methods for style transfer~\cite{gatys2016image, li2016combining,luan2017deep} using pre-trained feature extractors, such as VGG networks~\cite{Simonyan15}, can be divided into parametric and non-parametric methods. Categorized as parametric methods, some methods~\cite{gatys2016image, berger2016incorporating} designed Gram matrix to capture global statistics of features. However, the loss function based on Gram matrix leads to poor results as it captures the per-pixel feature correlations and does not constrain the spatial layout. To address this issue, non-parametric approaches~\cite{li2016combining,luan2017deep,aberman2018neural} match a style on patch-level. %However, they fail when a pair of images have significant discrepancies in the structure, 
Categorized as non-parametric methods, some works have realized style transfer inspired by an image analogy~\cite{hertzmann2001image}, which is based on dense correspondence~\cite{shih2014style,liao2017visual}. STROTSS~\cite{kolkin2019style} uses optimal transport algorithm. The aforementioned optimization-based methods are limited to encode an image translation prior, thus often generating artifacts and showing limited photorealism.

On the other hand, recent learning-based methods~\cite{chen2016fast,li2017universal,huang2017arbitrary,gu2018arbitrary,sanakoyeu2018style,li2018closed,yoo2019photorealistic,park2019arbitrary,liu2021adaattn} tried to solve style transfer by data-driven ways. They mostly focused on designing loss functions, and often included pre- or post-processing~\cite{li2018closed} to produce spatially smooth output. For instance, contextual loss is proposed~\cite{mechrez2018contextual}, which trains CNNs solely using the content images without need of large-scale paired dataset. Several works ~\cite{huang2017arbitrary, gu2018arbitrary,yoo2019photorealistic, qu2019one, park2019arbitrary} trained a decoder network with MS-COCO dataset~\cite{lin2014microsoft} or ImageNet dataset~\cite{deng2009imagenet}, or needed training the whole network per style~\cite{sanakoyeu2018style} before optimization process. However, they may be biased to the training images or styles and may not generalize well to unseen data.
%\textcolor[rgb]{1.0,0.0,0.0}{Both optimization- and learning-based methods are limited to encode a pair-specific image translation prior, thus often generating artifacts and showing limited photorealism.}

\paragraph{Image Prior.} 
Deep Image Prior (DIP)~\cite{ulyanov2018deep} proves the structure of generator network itself can serve as a prior for image restoration, against the assumption that learning from large-scale data is necessary to capture realistic image prior~\cite{zhang2018understanding}, of which many variants were proposed, tailored to
solve an inverse problem~\cite{burger2005nonlinear, dabov2007image,burger2012image}. SinGAN~\cite{shaham2019singan} and SinIR~\cite{yoo2021sinir} fine-tune GAN or AE on a single input and can be applied to image manipulation and restoration.
GAN inversion~\cite{gansteerability2020ja,menon2020pulse,gu2020image} aims at generating an image by solely optimizing a latent code of pre-trained GAN given a target image. 
Different from the aforementioned methods that attempts to learn an image prior, our framework is the first attempt to learn the image translation prior.
%\textcolor[rgb]{1.0,0.0,0.0}{The aforementioned methods concentrate to leverage prior in a single image. Unlike these methods, our framework attempts to learn image translation prior between a pair of images through optimizing the feature extractor and generation module at test-time, dubbed Deep Translation Prior(DTP).} %Different from the aforementioned methods, our framework is the first attempt to learn the image \emph{translation} prior.
\begin{figure*}[t]
\centering
\includegraphics[width=0.96\linewidth]{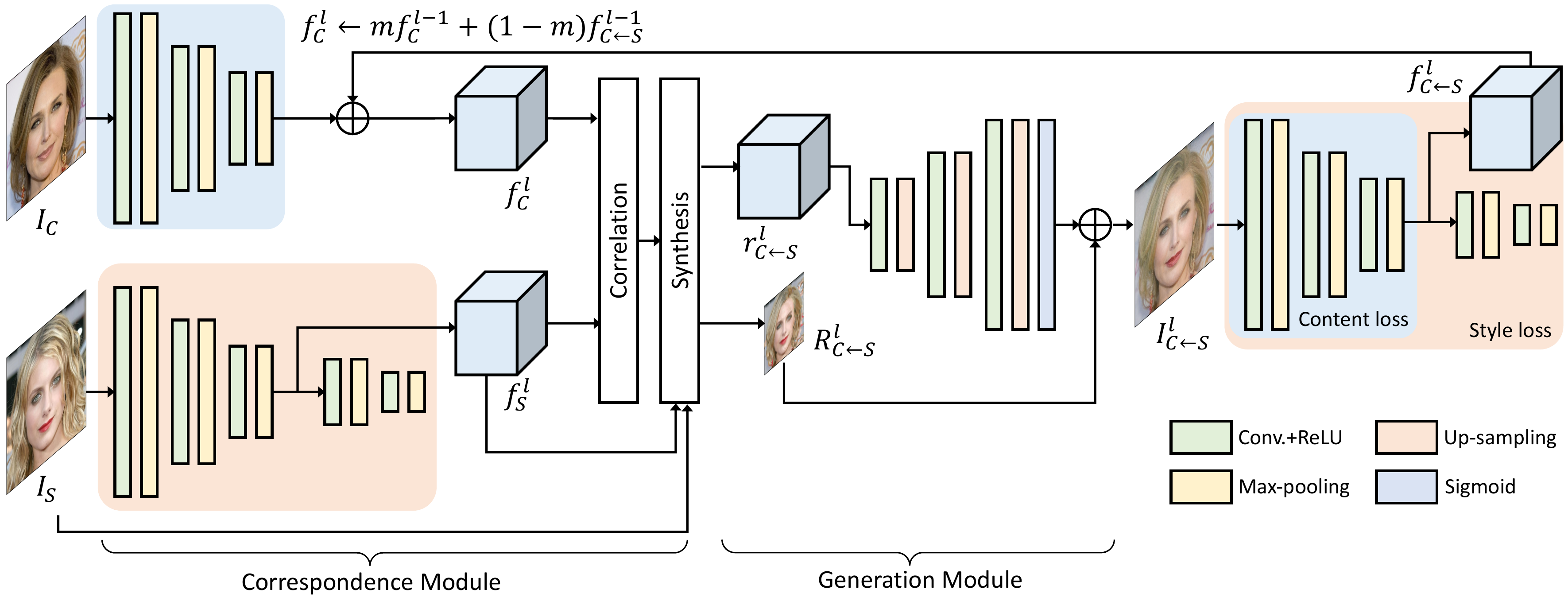}\\
  \caption{\textbf{Network configuration of DTP.} 
  Our network consists of two sub-modules, \emph{correspondence} module and \emph{generation} module. At the first, we predict a translation hypothesis by first computing the similarity between each source point and all target points and then warping the style image and feature in a probabilistic manner. At the second, the warped feature goes through a decoding network that generates the residual of the final stylized image.} 
\label{cft_overall}\vspace{-10pt}
\end{figure*}

\section{Methodology}
\subsection{Motivation}
Photorealistic style transfer aims at transferring the style of image $I_\mathcal{S}$ to the content of image $I_\mathcal{C}$ to synthesize a stylized image $I_{\mathcal{C} \leftarrow \mathcal{S}}$. To achieve this, traditional methods~\cite{li2016combining,gatys2016image,luan2017deep} focused on an image \emph{optimization} technique, from which deep convolutional features were extracted from content and style images, denoted by $F_\mathcal{C}=\Phi(I_\mathcal{C})$ and $F_\mathcal{S}=\Phi(I_\mathcal{S})$ with feature extractor $\Phi(\cdot)$, and used to define an objective function, consisting of content loss $\mathcal{L}_\mathrm{cont}$ and style loss $\mathcal{L}_\mathrm{style}$ functions, as in~\figref{img:2}(a):
\begin{equation}
I_{\mathcal{C} \leftarrow \mathcal{S}} =\underset{I}{\mathrm{argmin}}
\{\mathcal{L}_\mathrm{cont}(\Phi(I),F_\mathcal{C}) + \mathcal{L}_\mathrm{style}(\Phi(I),F_\mathcal{S})\}.
\end{equation}
Since it is often a non-convex optimization, most methods leverage an iterative solver, e.g., gradient descent~\cite{li2016combining,gatys2016image,luan2017deep}, and thus they benefit from an error feedback to find better stylized images. However, they are limited to encode an image \emph{translation} prior on synthesized images, and thus often generate artifacts and limited photorealism.

To overcome these limitations, recent \emph{learning}-based methods~\cite{li2017universal,huang2017arbitrary,li2018closed,gu2018arbitrary,yoo2019photorealistic} attempt to learn such translation prior within the networks during training. Starting with feature extraction module, they designed feature fusion module, e.g., AdaIN~\cite{huang2017arbitrary} or WCT~\cite{li2017universal}, and trained decoder module on large-scale image data, e.g., ImageNet~\cite{deng2009imagenet} or MS-COCO~\cite{lin2014microsoft}, as in~\figref{img:2}(b), which can be formulated as
\begin{equation}
\omega^\dagger =\underset{\omega}{\mathrm{argmin}} 
\sum\nolimits_{n}\mathcal{L}_{\mathrm{recon}}({\mathcal{F}}(F_{\mathcal{C},n},F_{\mathcal{S},n};\omega),I_{\mathcal{C} \leftarrow \mathcal{S},n}),
\end{equation}
where $F_{\mathcal{C},n}$ and $F_{\mathcal{S},n}$ are features from $n$-th image pair and $I_{\mathcal{C} \leftarrow \mathcal{S},n}$ is $n$-th stylized image sampled from massive training data. $\mathcal{F}(\cdot;\omega)$ is a feed-forward process with decoder parameters $\omega$. $\mathcal{L}_{\mathrm{recon}}$ is an image reconstruction loss function~\cite{gatys2016image,liu2017unsupervised,huang2018multimodal,park2019semantic}. In practice, since it is notoriously challenging to collect training pairs for style transfer, $\{(F_{\mathcal{C},n},F_{\mathcal{S},n},I_{\mathcal{C} \leftarrow \mathcal{S},n})\}_{n\in\{1,...,N\}}$, due to its subjectivity, most methods~\cite{li2017universal,huang2017arbitrary,li2018closed,gu2018arbitrary,yoo2019photorealistic} alternatively leverage an auto-encoding setting to learn the decoder with parameters $\omega$ to reconstruct an input image itself learned by $\mathcal{L}_{\mathrm{recon}}({\mathcal{F}}({\Phi(I)};\omega),I)$. At test-time, given $F_{\mathcal{C}}$ and $F_{\mathcal{S}}$, a stylization process can be formulated as follows:
\begin{equation}
I_{\mathcal{C} \leftarrow \mathcal{S}} = \mathcal{F}(F_\mathcal{C},F_\mathcal{S};\omega^\dagger).
\end{equation}
These methods are based on the assumption that the image translation prior can be learned within the model itself from massive training data. However, during the training phase, these methods do not leverage \emph{explicit} content and style loss functions, as done in optimization methods~\cite{li2016combining,gatys2016image,luan2017deep}, thus providing limited stylization performance when their assumptions are violated, e.g., under unseen images or styles. In addition, adopting fixed network parameters at test-time may not capture an image pair-specific translation prior.

\subsection{Overview}
To overcome aforementioned limitations and take the best of both approaches, we present Deep Translation Prior (DTP) framework. We argue that the translation prior does not necessarily need to be learned from intensive learning or datasets. Instead, an 
\emph{image pair-specific translation prior} can be captured by solely \emph{minimizing explicit content and style loss functions on the image pair}, like what is done by conventional optimization-based methods~\cite{li2016combining,gatys2016image,luan2017deep}, with an \emph{untrained network for stylization}, which takes benefits of large capacity and robustness of networks as in recent learning-based methods~\cite{li2017universal,huang2017arbitrary,li2018closed,gu2018arbitrary,yoo2019photorealistic}, as in~\figref{img:2}(c), formulated as
\begin{equation}
\begin{split}
&\omega^{*} =\underset{\omega}{\mathrm{argmin}}
\{\mathcal{L}_\mathrm{cont}(\Phi(\mathcal{F}(F_\mathcal{C},F_\mathcal{S};\omega)),F_\mathcal{C}) \\
& \qquad \qquad \qquad + \mathcal{L}_\mathrm{style}(\Phi(\mathcal{F}(F_\mathcal{C},F_\mathcal{S};\omega)),F_\mathcal{S})\}, \\
&I_{\mathcal{C} \leftarrow \mathcal{S}}=\mathcal{F}(F_\mathcal{C},F_\mathcal{S};\omega^{*}),
\end{split}
\end{equation}
where $\omega^{*}$ is overfitted to the input image pair, which encodes the image pair-specific translation prior. 
Unlike conventional optimization-based methods~\cite{gatys2016image,li2016combining}, our framework generates better stylization results while eliminating the artifacts thanks to the \emph{structure} of networks that can encode the image pair-specific prior during test-time training. In addition, unlike recent learning-based methods~\cite{li2017universal,huang2017arbitrary,li2018closed,gu2018arbitrary,yoo2019photorealistic}, our framework does not require an intensive training for decoder, but only requires an off-the-shelf feature extractor and untrained generator, thus having better generalization ability to unseen images or styles.

Tailored for such test-time training for style transfer, we design our stylization networks in a two-stage fashion, as illustrated in~\figref{cft_overall}; on one hand, the model predicts a translation hypothesis by first computing the similarity between each content point and all style points by means of the feature vectors $F_\mathcal{C}$ and $F_\mathcal{S}$, called \emph{correspondence} module, and on the other, the model refines the hypothesis through the decoder for more plausible stylization, called \emph{generation} module. 
Since the generated output is desired to preserve the structure of the content image $I_\mathcal{C}$ while faithfully stylizing from semantically similar parts in the style image $I_\mathcal{S}$, we present contrastive content loss function and style loss function to boost the convergence of our test-time training framework. 

It should be noted that there exist similar literature for image restoration tasks, e.g., Deep Image Prior (DIP)~\cite{ulyanov2018deep}, that have shown that the structure of a generator network can capture a low-level \emph{image} prior by optimizing a randomly-initialized network with a task-dependent fidelity term on a single image. To the best of our knowledge, our framework is the first attempt to learn the \emph{translation} prior at test-time for photorealistic style transfer.

\begin{figure}[t]
	\centering
	\renewcommand{\thesubfigure}{}
	\includegraphics[width=0.245\linewidth]{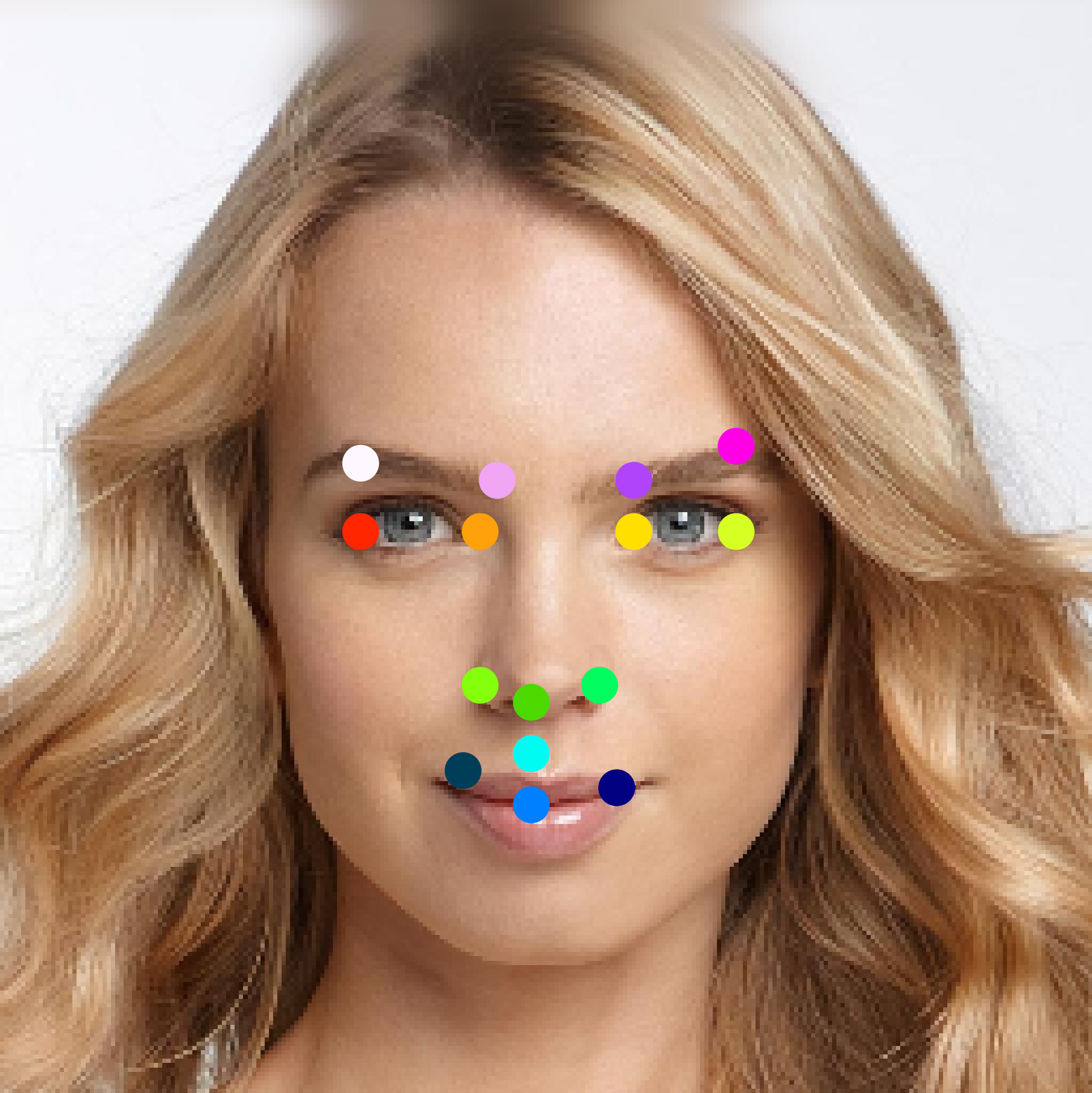}\hfill
	\includegraphics[width=0.245\linewidth]{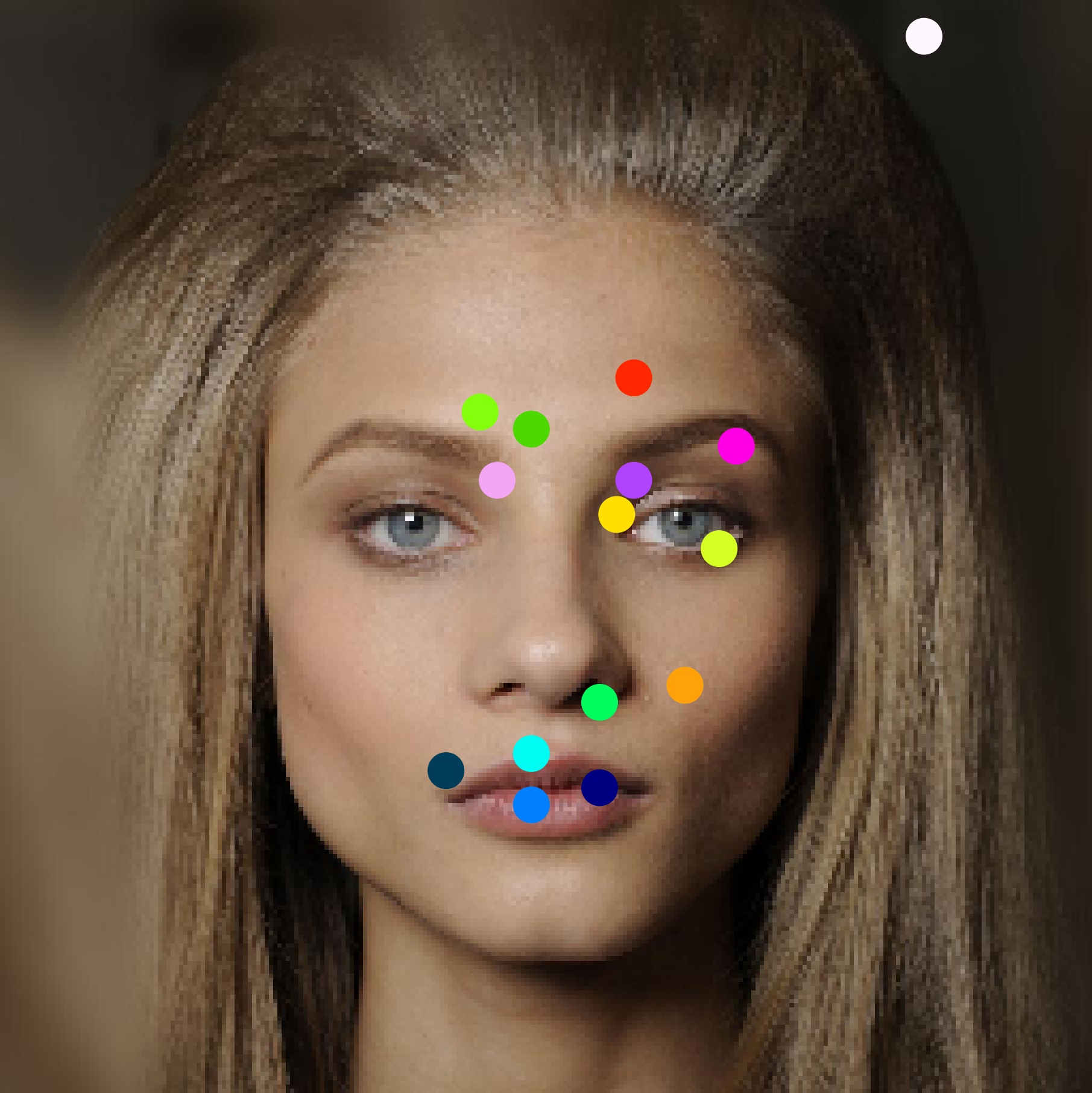}\hfill
	\includegraphics[width=0.245\linewidth]{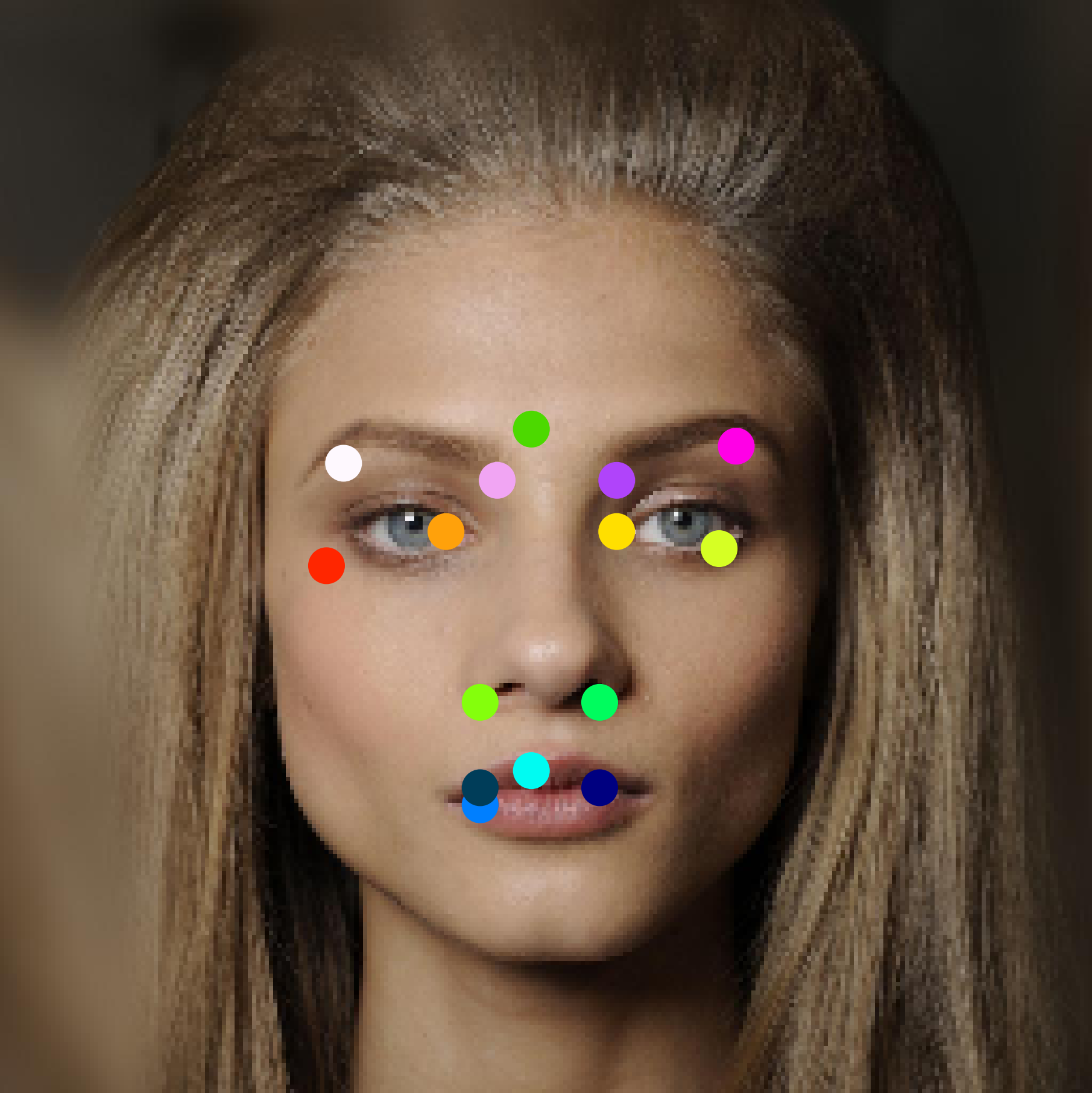}\hfill
	\includegraphics[width=0.245\linewidth]{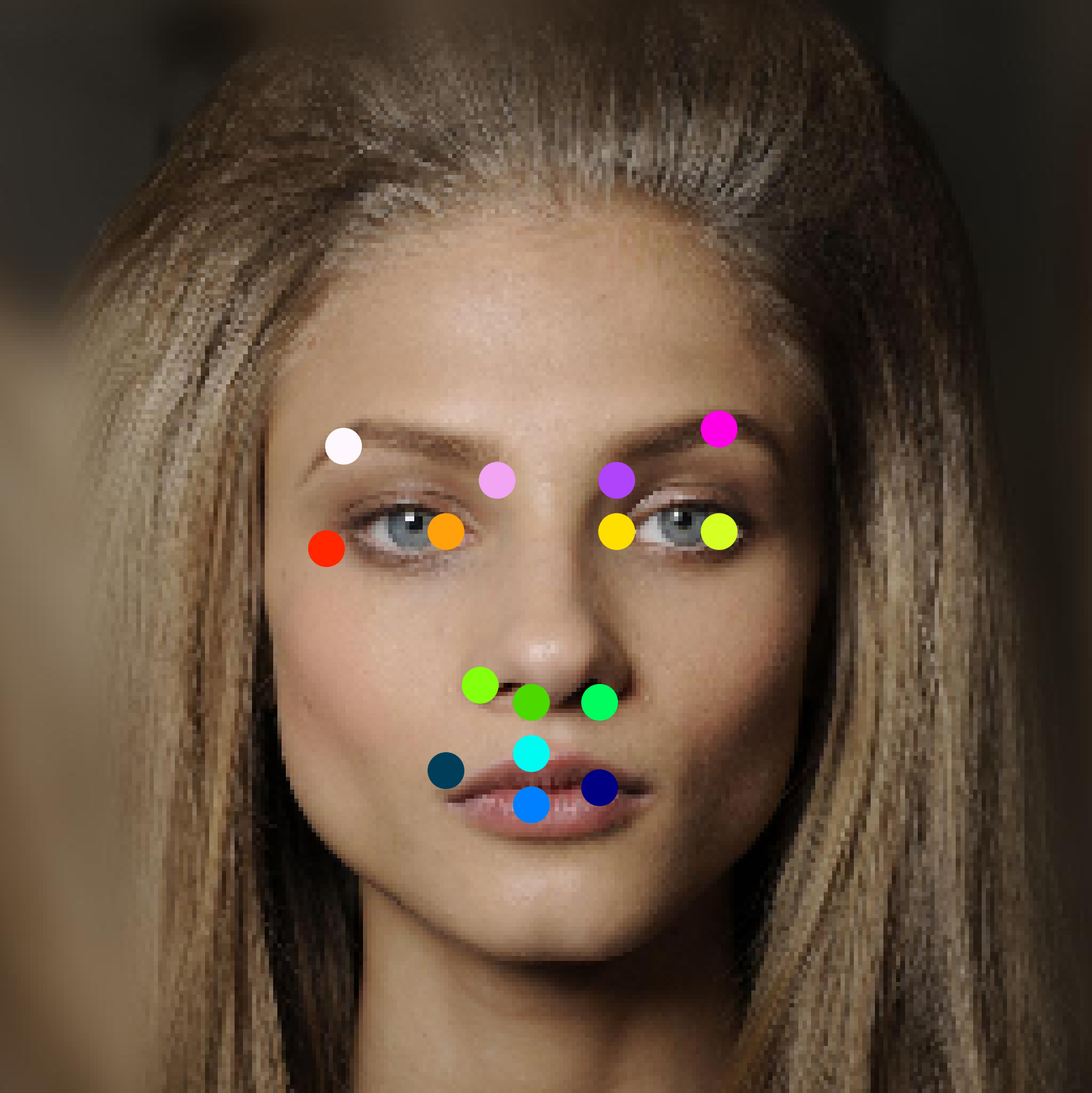}\hfill\\
	\vspace{-5pt}
  \caption{\textbf{Visualization of matched points:} (from left to right) content image with query points, and style images with initial, intermediate, and final matched points.}\vspace{-10pt}\label{img:matching}
\end{figure}

\subsection{Network Architecture}
\paragraph{Correspondence Module.}\label{sec:3_2}
We first present a correspondence module to measure the similarities between each point in content feature $f_\mathcal{C}$ and all other points in style feature $f_\mathcal{S}$, enabling generating a translation hypothesis. 
It is inspired by the classical matching pipeline~\cite{rocco2017convolutional} in that we first extract the feature vectors and then compute the similarity between them.

Following the previous approaches for style transfer, 
we first extract the deep convolution features, e.g., VGGNet~\cite{Simonyan15} pretrained on ImageNet~\cite{deng2009imagenet}, as follows:
\begin{equation}
\begin{split}
    & f_\mathcal{C} = \Phi(I_\mathcal{C};\omega_f) \in \mathbb{R}^{H\times W\times C},\\
    & f_\mathcal{S} = \Phi(I_\mathcal{S};\omega_f) \in \mathbb{R}^{H\times W\times C}, 
\end{split}
\end{equation}
where $H$ and $W$ are spatial size, with $C$ channels of $f$. $\omega_f$ are feature extraction parameters. Unlike most existing methods that use \emph{fixed} feature extraction parameters, we adaptively \emph{fine-tune} the parameters to the input image pair. We then compute a correlation matrix $M \in \mathbb{R}^{HW \times HW}$, of which each term is a pairwise feature correlation such that
\begin{equation}
    M(u,v) = \frac{\hat{f}_\mathcal{C}(u)^T \hat{f}_\mathcal{S}(v)}{\|\hat{f}_\mathcal{C}(u)\|\|\hat{f}_\mathcal{S}(v)\|},
\end{equation}
where $u$ and $v$ represent all the points in the content and style images, respectively. $\hat{f}_\mathcal{C}(u)$ and $\hat{f}_\mathcal{S}(v)$ are channel-wise centralized features of $f_\mathcal{C}(u)$ and $f_\mathcal{S}(v)$ as
\begin{equation}
    \hat{f}_\mathcal{C}(u) = f_\mathcal{C}(u) - \Bar{f_\mathcal{C}},
    \quad \hat{f}_\mathcal{S}(v) = f_\mathcal{S}(v) - \Bar{f_\mathcal{S}}, 
\end{equation}
where $\Bar{f_\mathcal{C}}$ is an average of $f_\mathcal{C}(u)$ across all the points in the content. $\Bar{f_\mathcal{S}}$ is similarly defined.
Since $M(u,v)$ represents a similarity between $u$ and $v$, the higher, the more similar.

By using the correlation matrix $M$, we synthesize an warped style feature $r_{\mathcal{C} \leftarrow \mathcal{S}}$, i.e., the style feature spatially-aligned to the content image. The warping function can be formulated in many possible ways, but we borrow the technique in~\cite{zhang2020cross} that uses a reconstruction:
\begin{equation}
    r_{\mathcal{C} \leftarrow \mathcal{S}}(u) = \sum\nolimits_{v}^{} \Omega (M(u,v)/\tau) f_\mathcal{S}(v), 
\end{equation}
where $\Omega$ means the softmax operator across $v$, and $\tau$ is a temperature parameter. Matched points in our framework is visualized in~\figref{img:matching}.

\paragraph{Generation Module.}\label{sec:3_3}
Our generation module aims at reconstructing an image from warped feature $r_{\mathcal{C} \leftarrow \mathcal{S}}$. We present the decoder that has a symmetric structure of feature extractor architecture, similar to~\cite{li2017universal, huang2017arbitrary}. This decoding process can be formulated as follows:
\begin{equation}
    I_{\mathcal{C} \leftarrow \mathcal{S}} = \mathcal{F}(r_{\mathcal{C} \leftarrow \mathcal{S}};\omega_{g}),
\end{equation}
where $\omega_{g}$ is decoding parameters. As described above, the parameters are first \emph{randomly}-initialized and then learned with \emph{explicit} loss functions for style transfer at test time. 

However, due to non-convexity of the loss functions for style transfer, generating the image $I_{\mathcal{C} \leftarrow \mathcal{S}}$ through the decoder directly is extremely hard to converge. To elevate the stability and boost the convergence, we exploit not only warped style feature $r_{\mathcal{C} \leftarrow \mathcal{S}}$, but also warped style image $R_{\mathcal{C} \leftarrow \mathcal{S}}$, extracted such that $R_{\mathcal{C} \leftarrow \mathcal{S}}(u) = \sum_{v}^{} {\Omega} (M(u,v)/\tau) I_\mathcal{S}(v)$, as a guidance for style transfer, where the networks only learn the residual for the final result as follows:
\begin{equation}
    I_{\mathcal{C} \leftarrow \mathcal{S}} = \lambda_{w} \mathcal{F}(r_{\mathcal{C} \leftarrow \mathcal{S}};\omega_{g}) + (1-\lambda_{w}) R_{\mathcal{C \leftarrow S}},
\end{equation}
where $\lambda_{w}$ is a weight parameter. By leveraging such a residual prediction, convergence of our test-time training could be greatly improved. Moreover, it enables directly flowing the loss gradients to both feature extractor with $\omega_{f}$ and image generator with $\omega_{g}$, which helps to boost the performance.
\begin{figure*}[t]
\centering
\includegraphics[width=0.123\linewidth]{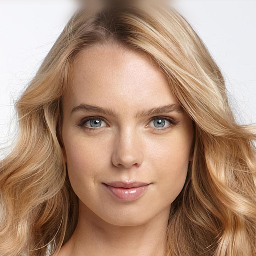}\hfill
\includegraphics[width=0.123\linewidth]{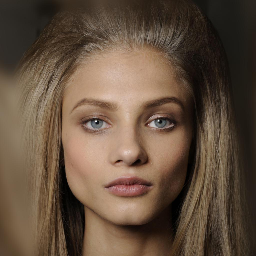}\hfill
\includegraphics[width=0.123\linewidth]{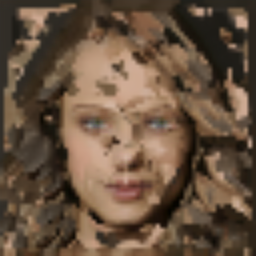}\hfill
\includegraphics[width=0.123\linewidth]{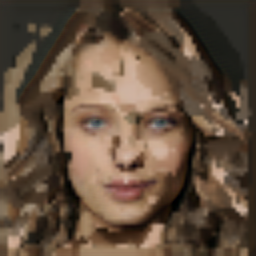}\hfill
\includegraphics[width=0.123\linewidth]{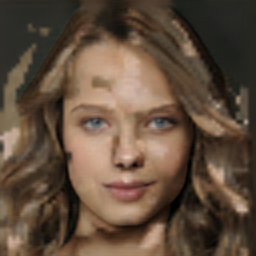}\hfill
\includegraphics[width=0.123\linewidth]{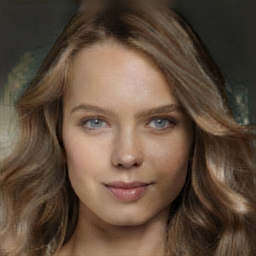}\hfill
\includegraphics[width=0.123\linewidth]{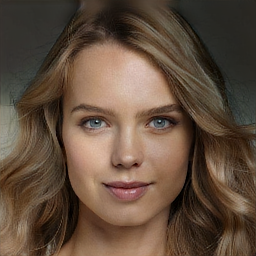}\hfill
\includegraphics[width=0.123\linewidth]{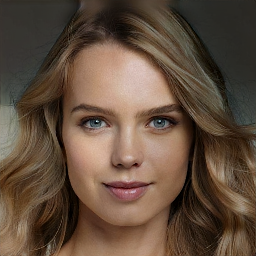}\hfill
\\ \vspace{-5pt}
  \caption{\textbf{DTP optimization:} (from left to right) content and style images, and stylized images as evolving an iteration, which shows how the optimization converges.}
\label{iteration}\vspace{-10pt}
\end{figure*}

\paragraph{Iterative Formulation.}\label{sec:3_4}
Since the loss function for style transfer, which will be discussed later, is non-convex, we formulate our test-time training as an iterative framework, as exemplified in~\figref{iteration}. 
As evolving the iteration, the image $I^l_{\mathcal{C} \leftarrow \mathcal{S}}$ at $l$-th iteration converges to better stylization results, since it is generated from the updated feature extractor parameters $\omega_{f}$ and decoder parameters $\omega_{g}$. Since the image $I^l_{\mathcal{C} \leftarrow \mathcal{S}}$ is getting close to the optimal, if the content image $I_\mathcal{C}$ can be substituted by $I^l_{\mathcal{C} \leftarrow \mathcal{S}}$ in a recurrent fashion, the iterative solver can converge faster and boost performance. 
However, at early stages during optimization, $I^l_{\mathcal{C} \leftarrow \mathcal{S}}$ contains blurry regions and noises, which prohibit using such an explicit recurrent formulation. 
To overcome this limitation, we adopt a moving averaging technique similar to~\cite{kim2019semantic,schmidt2020towards} in a manner that we smoothly substitute the content feature $f_\mathcal{C}$ by the output feature $f_{\mathcal{C} \leftarrow \mathcal{S}} = \Phi(I_{\mathcal{C} \leftarrow \mathcal{S}};\omega_{f})$ with a momentum parameter $m$, such that
\begin{equation}
    f^l_\mathcal{C} \leftarrow m f^{l}_\mathcal{C} + (1-m) f^{l-1}_{\mathcal{C} \leftarrow \mathcal{S}},
\end{equation}
which is used to the current content feature. 
\begin{figure*}[t]
	\centering
	\renewcommand{\thesubfigure}{}
	\subfigure[]
	{\includegraphics[width=0.122\linewidth]{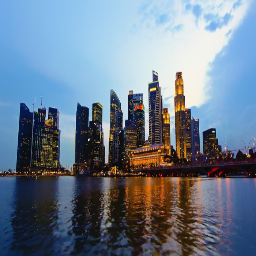}}\hfill
	\subfigure[]
	{\includegraphics[width=0.122\linewidth]{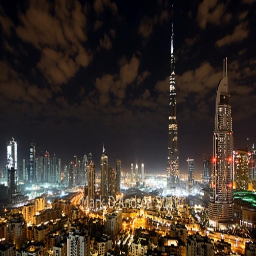}}\hfill
	\subfigure[]
	{\includegraphics[width=0.122\linewidth]{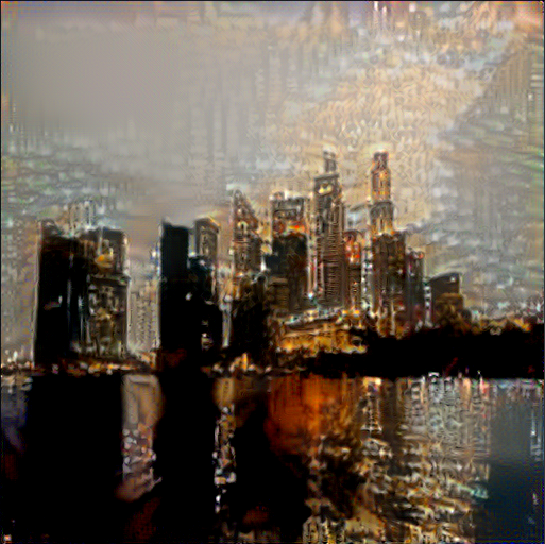}}\hfill
	\subfigure[]
	{\includegraphics[width=0.122\linewidth]{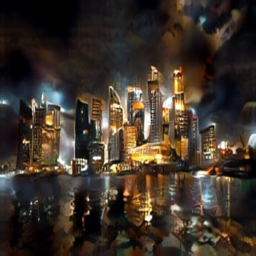}}\hfill
	\subfigure[]
	{\includegraphics[width=0.122\linewidth]{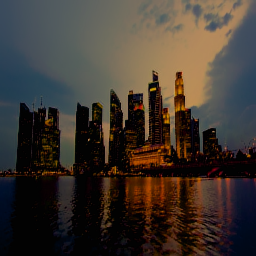}}\hfill
	\subfigure[]
	{\includegraphics[width=0.122\linewidth]{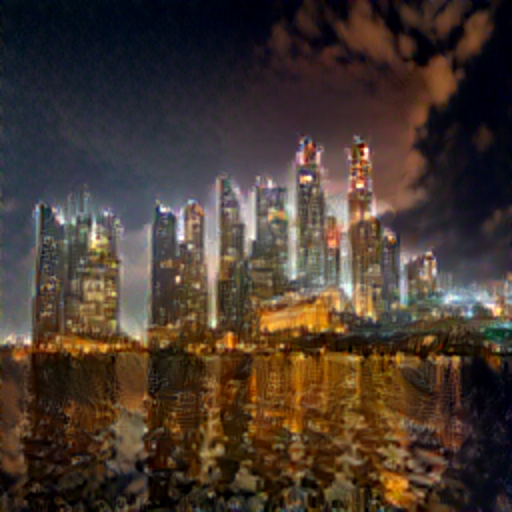}}\hfill
	\subfigure[]
	{\includegraphics[width=0.122\linewidth]{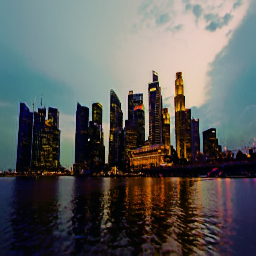}}\hfill	
	\subfigure[]
	{\includegraphics[width=0.122\linewidth]{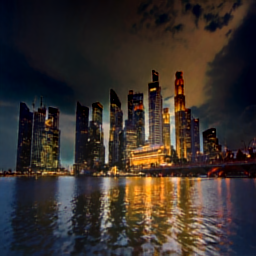}}\hfill\\
	\vspace{-20.5pt}	
	\subfigure[Content]
	{\includegraphics[width=0.122\linewidth]{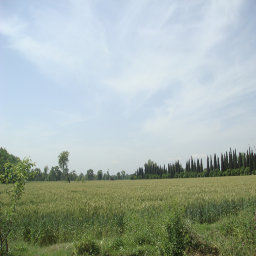}}\hfill
	\subfigure[Style]
	{\includegraphics[width=0.122\linewidth]{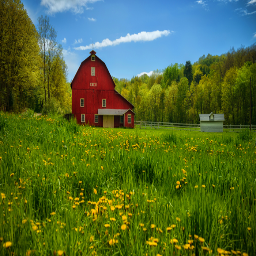}}\hfill
	\subfigure[(a)]
	{\includegraphics[width=0.122\linewidth]{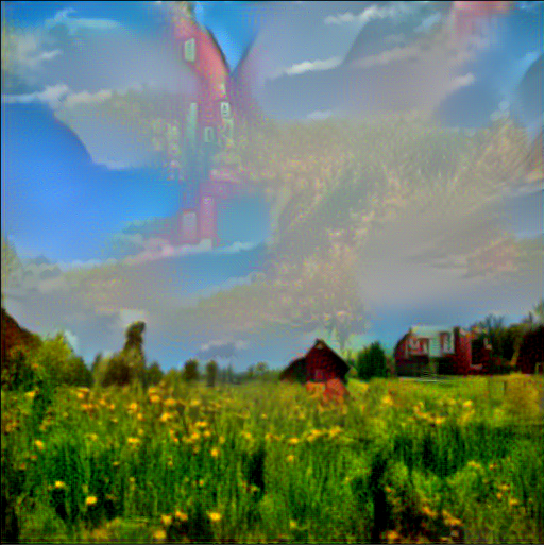}}\hfill
	\subfigure[(b)]
	{\includegraphics[width=0.122\linewidth]{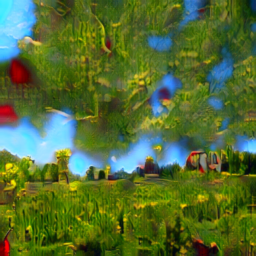}}\hfill
	\subfigure[(c)]
	{\includegraphics[width=0.122\linewidth]{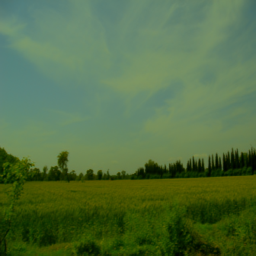}}\hfill
	\subfigure[(d)]
	{\includegraphics[width=0.122\linewidth]{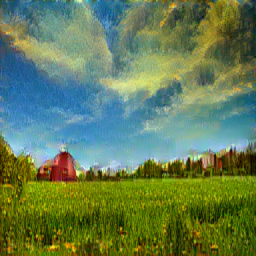}}\hfill
	\subfigure[(e)]
	{\includegraphics[width=0.122\linewidth]{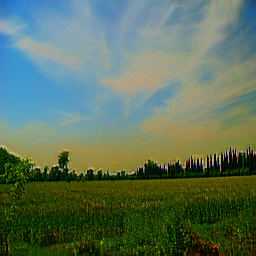}}\hfill	
	\subfigure[(f)]
	{\includegraphics[width=0.122\linewidth]{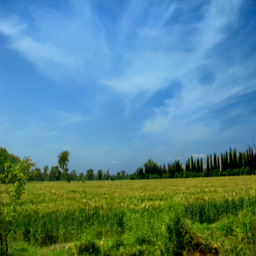}}\hfill\\
	\vspace{-10pt}
    \caption{\textbf{Comparison of DTP with other methods on standard benchmark~\cite{luan2017deep,an2019ultrafast}:} Given content and style images, stylized results are achieved by (a) Gatys et al.~\cite{gatys2016image}, (b) WCT~\cite{li2017universal}, (c) WCT2~\cite{yoo2019photorealistic}, (d) STROTSS~\cite{kolkin2019style}, (e) SinIR~\cite{yoo2021sinir}, and (f) Ours. 
	Compared to others, our model generates realistic results while successfully transferring both global and local style information and preserving structure. Our model does not require any mask, pre- or post-processing.}
	 \label{photo_result}\vspace{-5pt}
\end{figure*}

\subsection{Loss functions}\label{sec:3_5}
\paragraph{Content Loss.}
In photorealistic style transfer, the structure of content image should be preserved on the output image, and thus the content loss is generally defined as the feature difference between the content image $I_\mathcal{C}$ and the generated image $I_{\mathcal{C} \leftarrow \mathcal{S}}$, e.g., $\|f_\mathcal{C}-f_{\mathcal{C} \leftarrow \mathcal{S}}\|^2$. However, this content loss does not consider other pixels, thus making the result blurry and inducing the trivial solution when the features are simultaneously trained as in our test-time training. 
To overcome this, we revisit infoNCE loss~\cite{oord2018representation} to set the \emph{pseudo} positive samples between content image $I_\mathcal{C}$ and generated image $I_{\mathcal{C} \leftarrow \mathcal{S}}$. We first encode the feature stacks as used in \cite{chen2020simple, park2020contrastive}, compiling stacks of features $\{f^l_\mathcal{C}\}$ and $\{f^l_{\mathcal{C} \leftarrow \mathcal{S}}\}$, where $l \in \{1, ...,L_\mathcal{C}\}$. We define the exponential of inner product function of two vectors to express the equation more comfortably.
\begin{equation}
s(f,g) = \exp
\left(\left(f^T g / \|f\|\|g\|\right)/ \tau\right),
\end{equation}
where $\tau$ is a temperature parameter. Our final content loss is then defined as follows:
\begin{equation}
    \mathcal{L}_\mathrm{cont} = -\sum_{l}\sum_{u}\mathrm{log}
    \left(\frac{s(f^l_\mathcal{C}(u),f^l_{\mathcal{C} \leftarrow \mathcal{S}}(u))}
    {\sum_{v} s(f^l_\mathcal{C}(u), f^l_{\mathcal{C} \leftarrow \mathcal{S}}(u)) }\right).
\end{equation} 

\paragraph{Style Loss.}
%While utilizing $\mathcal{L}_\mathrm{cont}$ helps to reconstruct the translated image $I_{\mathcal{C} \leftarrow \mathcal{S}}$ having content structure of $I_\mathcal{C}$, but may lose some style information of style image $I_\mathcal{S}$. To overcome this, 
We additionally adopt style loss functions. Unlike parametric methods~\cite{gatys2016image,li2017universal,chen2017photographic,li2018closed,yoo2019photorealistic} that enforce the style loss globally, we adopt the style loss similar to non-parametric methods~\cite{li2016combining,kim2019semantic} for getting detailed results which are more suitable for photo realistic style transfer. Similar to the content loss, the style loss is defined in a multi-scale manner. 
Our style loss is defined as below:
\begin{equation}
    \mathcal{L}_\mathrm{style} = \sum\nolimits_{l} \sum\nolimits_{v} \| \Psi(f^l_{\mathcal{C} \leftarrow \mathcal{S}}(v)) - \Psi(f^l_{\mathcal{S}}({NN(v)})) \|_{F}^{2},
\end{equation}
for $l \in \{1, ...,L_\mathcal{S}\}$. $\|\cdot\|_{F}^{2}$ denotes a Frobenius norm. $NN(v)$ is the index of the patch in $\Psi(f_\mathcal{S})$ that is the nearest patch of $\Psi(f_{\mathcal{C} \leftarrow \mathcal{S}}(v))$. 
\begin{figure*}[t]
	\centering
	\renewcommand{\thesubfigure}{}
	\subfigure[]
	{\includegraphics[width=0.122\linewidth]{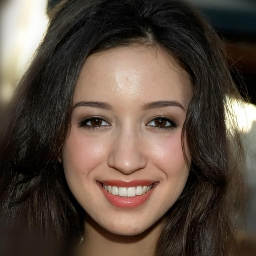}}\hfill
	\subfigure[]
	{\includegraphics[width=0.122\linewidth]{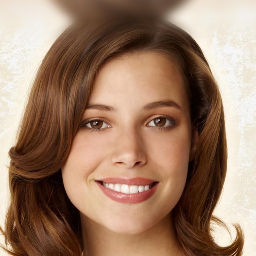}}\hfill
	\subfigure[]
	{\includegraphics[width=0.122\linewidth]{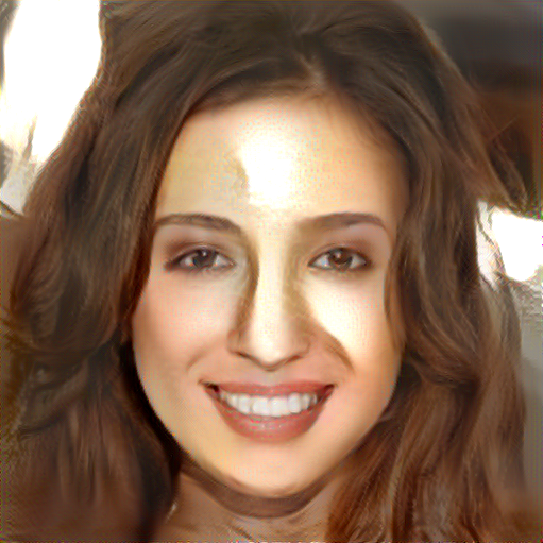}}\hfill
	\subfigure[]
	{\includegraphics[width=0.122\linewidth]{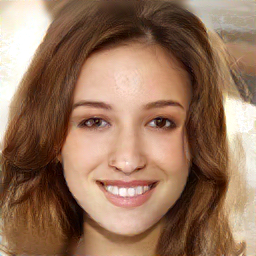}}\hfill
	\subfigure[]
	{\includegraphics[width=0.122\linewidth]{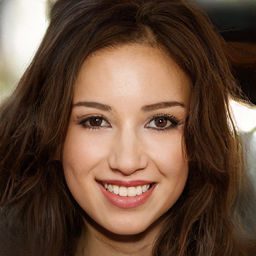}}\hfill
	\subfigure[]
	{\includegraphics[width=0.122\linewidth]{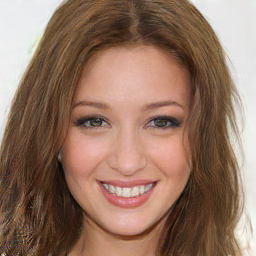}}\hfill
	\subfigure[]
	{\includegraphics[width=0.122\linewidth]{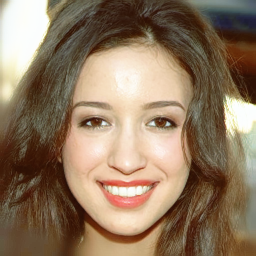}}\hfill
	\subfigure[]
	{\includegraphics[width=0.122\linewidth]{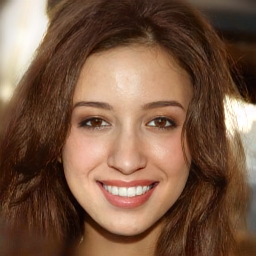}}\hfill\\
    \vspace{-20.5pt}
    \subfigure[Content]
	{\includegraphics[width=0.122\linewidth]{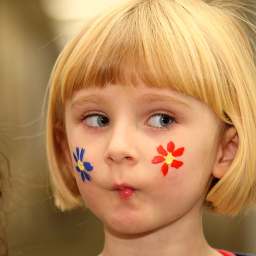}}\hfill
	\subfigure[Style]
	{\includegraphics[width=0.122\linewidth]{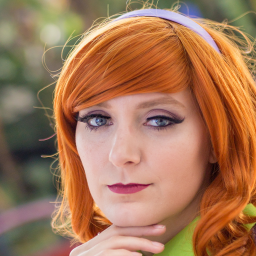}}\hfill
	\subfigure[ (a) ]
	{\includegraphics[width=0.122\linewidth]{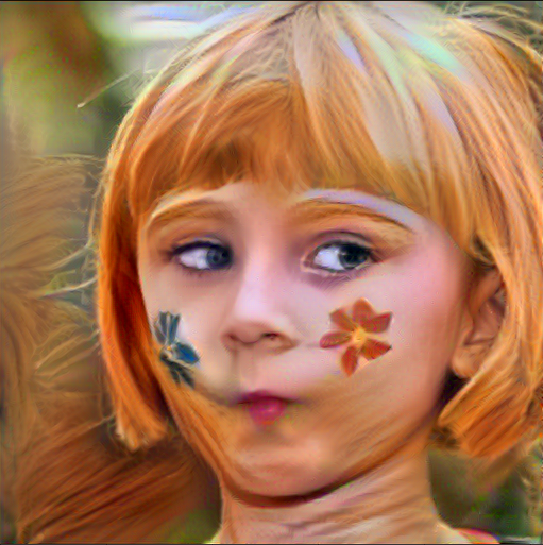}}\hfill
	\subfigure[ (b) ]
	{\includegraphics[width=0.122\linewidth]{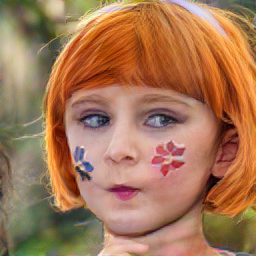}}\hfill
	\subfigure[ (c) ]
	{\includegraphics[width=0.122\linewidth]{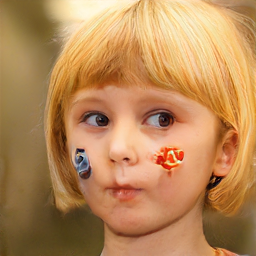}}\hfill
	\subfigure[ (d) ]
	{\includegraphics[width=0.122\linewidth]{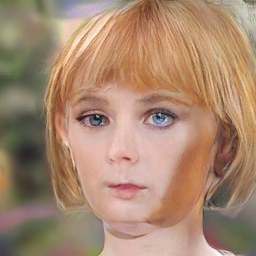}}\hfill
	\subfigure[ (e) ]
	{\includegraphics[width=0.122\linewidth]{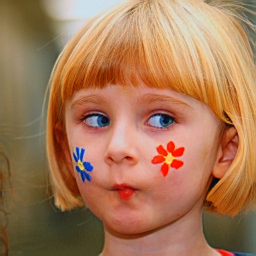}}\hfill
	\subfigure[ (f) ]
	{\includegraphics[width=0.122\linewidth]{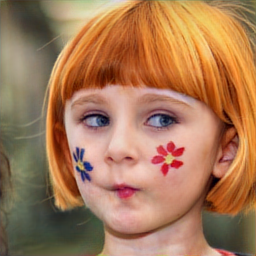}}\hfill
	\vspace{-10pt}
	\caption{\textbf{Comparison of DTP with other methods on CelebA-HQ dataset~\cite{liu2015deep} and FFHQ dataset~\cite{karras2019style}:} Given content and style images, translation results are achieved by (a) Gatys et al.~\cite{gatys2016image}, (b) STROTSS~\cite{kolkin2019style}, (c) Swapping Autoencoder~\cite{park2020swapping}, (d) CoCosNet~\cite{zhang2020cross}, (e) SinIR~\cite{yoo2021sinir}, and (f) Ours. The results show that our networks can translate local features as well as global features from both structure and style. Note that CoCosNet~\cite{zhang2020cross} was trained on CelebA-HQ, including segmentation masks, and Swapping Autoencoder~\cite{park2020swapping} was trained on FFHQ.}
	\label{celebahq_result}\vspace{-10pt}
\end{figure*}

\paragraph{Cycle Consistency Loss.}
To improve the stability during training, we further present cycle consistency loss as a regularization which enforces the stylized feature $r$ should be back-warped to the original feature $f$ well, defined such that
\begin{equation}
    r_{\mathcal{S}} = 
    \sum_{u}^{} \Omega \left(\frac{M(u,v)}{\tau}\right)
    \sum_{v}^{} \Omega \left(\frac{M(u,v)}{\tau}\right)  f_\mathcal{S}(v). 
\end{equation}
$r_{\mathcal{C}}$ is similarly defined. Then the cycle consistency loss $\mathcal{L}_\mathrm{cyc}$ is bidirectionally defined as
\begin{equation}
    \mathcal{L}_\mathrm{cyc} = 
    \sum_{u} \{
    \| f_{\mathcal{C}}(u) - r_{\mathcal{C}}(u)  \|_{F}^{2} 
    + \| f_{\mathcal{S}}(u) - r_{\mathcal{S}}(u) \|_{F}^{2} \}. 
\end{equation}

\paragraph{Total Loss.}
Finally, the total loss can be summarized such that $\mathcal{L} = \lambda_\mathrm{c}\mathcal{L}_\mathrm{cont} + (1-\lambda_\mathrm{c})\mathcal{L}_\mathrm{style} + \lambda_\mathrm{cyc}\mathcal{L}_\mathrm{cyc}$, where $\lambda_{c}$ and $\lambda_\mathrm{cyc}$ represent loss adjusting hyperparameters. 

\section{Experiments}
\subsection{Implementation Details}
We first summarize implementation details in our framework. For feature extractor, we used the ImageNet~\cite{deng2009imagenet} pre-trained VGG-19~\cite{Simonyan15} network. We set the temperature parameter $\tau$ as $0.07$ and weight parameter $\lambda_{w}$ as $1/9$. 
We also empirically set momentum parameter $m$ as $0.4$. $\lambda_\mathrm{c}=1/5$ and $\lambda_\mathrm{cyc}=1$ were used to adjust loss functions. We set the learning rates as $1e^{-4}$. We conduct experiments using a single 24GB RTX 3090 GPU. The network occupies memories about 6GB. We optimize our network over 1000 iterations, which takes about 150 seconds. The pair of content and style images are bilinearly resized to the size of 256$\times$256 in our experiment.

\subsection{Experimental Setup}
We used three kinds of datasets to evaluate our method, including standard datasets for photorealistic style transfer~\cite{luan2017deep,an2019ultrafast}, CelebA-HQ~\cite{liu2015deep}, and Flickr Faces HQ (FFHQ)~\cite{karras2019style}. We compared our method with recent state-of-the-art style transfer methods, such as Gatys et al.~\cite{gatys2016image}, WCT~\cite{li2017universal}, WCT2~\cite{yoo2019photorealistic}, STROTSS~\cite{kolkin2019style}, SinIR~\cite{yoo2021sinir}. We also compared with image-to-image translation tasks, such as CoCosNet~\cite{zhang2020cross}, and Swapping Autoencoder~\cite{park2020swapping}.
It should be emphasized that learning-based style transfer methods~\cite{li2017universal, yoo2019photorealistic} and image-to-image translation methods~\cite{zhang2020cross, park2020swapping} are trained on tremendous training data, while our method just trains the networks at test-time with a pair of images. \vspace{-5pt}
\begin{table}[t]
    \centering
    \scalebox{0.74}{
	\begin{tabular}{ 
	            >{\raggedright}m{0.28\linewidth}
				>{\centering}m{0.075\linewidth} 
				>{\centering}m{0.075\linewidth}
				>{\centering}m{0.075\linewidth} 
				>{\centering}m{0.075\linewidth}
				>{\centering}m{0.075\linewidth}
				>{\centering}m{0.075\linewidth}
				>{\centering}m{0.075\linewidth}
				>{\centering}m{0.075\linewidth}
				}
    \hlinewd{0.8pt}
    {}  & \multicolumn{2}{c}{Photo-R.} & \multicolumn{2}{c}{CelebA-HQ} & \multicolumn{2}{c}{FFHQ} & \multicolumn{2}{c}{Avg.} \tabularnewline
    \cline{2-9}
    Methods& PE$\downarrow$ &SSIM$\uparrow$  & PE$\downarrow$&SSIM$\uparrow$  &   PE$\downarrow$&SSIM$\uparrow$ &PE$\downarrow$&SSIM$\uparrow$ \tabularnewline
    \hline
    {Gatys et al.} & 3.09 & 0.44 & 4.37 & 0.66 & 1.97 & 0.69& 3.14 & 0.59  \tabularnewline
    {WCT} & 2.48 & 0.22 & - & - & - & - & - & - \tabularnewline
    {WCT2} & 1.99 & 0.70 & - & - & - & -& - & -  \tabularnewline
    {STROTSS} & 3.23 & 0.37 & 1.81 & 0.60 & 1.65 & 0.66& 1.88 & 0.54  \tabularnewline
    {Swap. AE} & 2.92 & 0.22 & 0.89 & 0.60 & {\bf 1.13} & 0.58 & 1.65 & 0.46  \tabularnewline
    {CoCosNet} & - & - & 1.56 & 0.43 & - & - & - & -  \tabularnewline
    
    {SinIR} & {\bf1.00} & {\bf 0.84} & 1.56  & 0.84 & 1.28 & 0.93 & 1.39 & 0.87  \tabularnewline
    \hline
%    DTP & 1.36 & 0.75 & {\bf 0.79} & {\bf 0.86} & 1.4 & {\bf 0.93}& {\bf 1.18} &  0.85  \tabularnewline
    DTP ($\lambda_\mathrm{c}=4/5$) & 1.36 & 0.75 & 0.79 & 0.86 & 1.4 & 0.93& 1.18 &  0.85  \tabularnewline
    DTP ($\lambda_\mathrm{c}=1/5$)  & 1.09 & 0.82 & {\bf 0.47} & {\bf 0.96} & 1.21 & {\bf 0.98}& {\bf 0.93} &  {\bf 0.92}  \tabularnewline
    \hlinewd{0.8pt}
    \end{tabular}}\vspace{-5pt}
    \caption{\textbf{Quantitative evaluation on standard benchmark} for photorealistic style transfer~\cite{luan2017deep,an2019ultrafast}, CelebA-HQ~\cite{liu2015deep}, and FFHQ~\cite{karras2019style}.}
    \label{quantitative_dtp}\vspace{-10pt}
\end{table}

\begin{figure*}[t]
\centering
	\renewcommand{\thesubfigure}{}
	\subfigure[]
	{\includegraphics[width=0.122\linewidth]{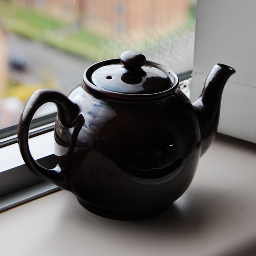}}\hfill
	\subfigure[]
	{\includegraphics[width=0.122\linewidth]{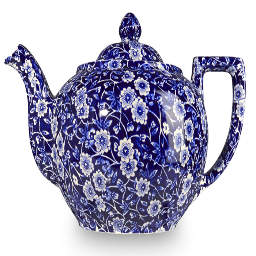}}\hfill
	\subfigure[]
	{\includegraphics[width=0.122\linewidth]{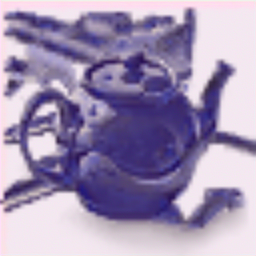}}\hfill
	\subfigure[]
	{\includegraphics[width=0.122\linewidth]{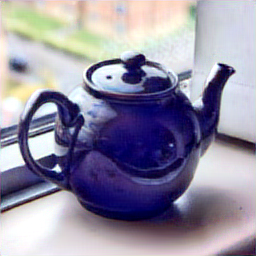}}\hfill
	\subfigure[]
	{\includegraphics[width=0.122\linewidth]{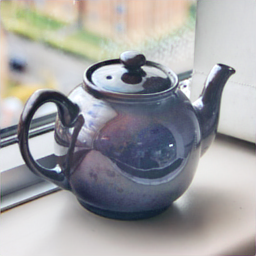}}\hfill
	\subfigure[]
	{\includegraphics[width=0.122\linewidth]{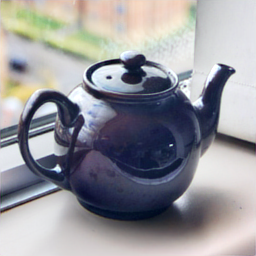}}\hfill
    \subfigure[]
	{\includegraphics[width=0.122\linewidth]{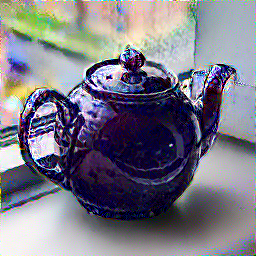}}\hfill
	\subfigure[]
	{\includegraphics[width=0.122\linewidth]{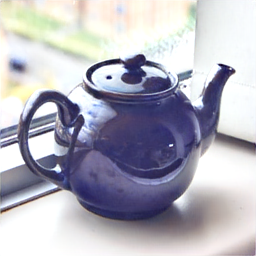}}\hfill\\
	\vspace{-20pt}	

	\subfigure[Content]
	{\includegraphics[width=0.122\linewidth]{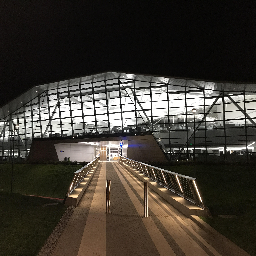}}\hfill
	\subfigure[Style]
	{\includegraphics[width=0.122\linewidth]{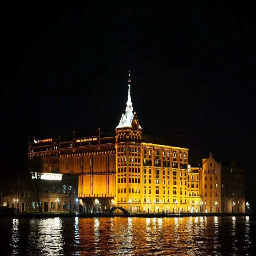}}\hfill
	\subfigure[w/o WF]	{\includegraphics[width=0.122\linewidth]{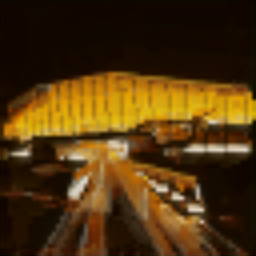}}\hfill
	\subfigure[w/o WI]
	{\includegraphics[width=0.122\linewidth]{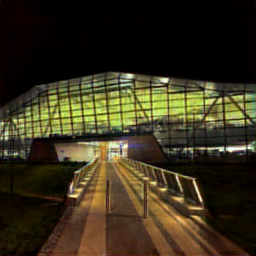}}\hfill
	%\subfigure[w/${P}_\mathrm{all}$]
	%{\includegraphics[width=0.1223\linewidth]{figure/ablation/64/result_woc_wom.png}}\hfill
	\subfigure[w/o FMA]
	{\includegraphics[width=0.122\linewidth]{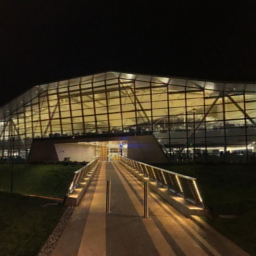}}\hfill
	\subfigure[w/o $\mathcal{L}_\mathrm{cyc}$]
	{\includegraphics[width=0.122\linewidth]{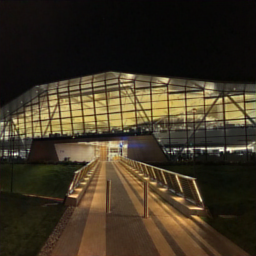}}\hfill
	\subfigure[w/o G. Module.]
	{\includegraphics[width=0.122\linewidth]{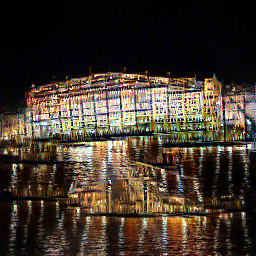}}\hfill
	\subfigure[Ours]
	{\includegraphics[width=0.122\linewidth]{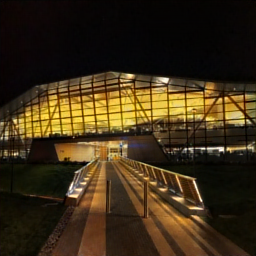}}\hfill\\
	\vspace{-10pt}
    \caption{\textbf{Ablation study on priors, cycle consistency loss, generation module and feature moving average:} Warped features (WF) and images (WI) help generation module to contain translation prior in the feature and image level. Feature moving average (FMA) helps to converge to better solution and $\mathcal{L}_\mathrm{cyc}$ stabilizes test-time optimization. Generation module (G. Module) helps to learn translation prior better.}
	\label{ablation_mov}\vspace{-10pt}
\end{figure*}

\subsection{Experimental Results}
\paragraph{Qualitative Evaluation.}
In this section, we evaluated photorealistic style transfer results of our method compared with state-of-the-art methods, with respect to two aspects, including synthesized image quality and semantic consistency. Qualitative results are shown in~\figref{photo_result} and \figref{celebahq_result}. Our generated results are realistic and contain both global-local style features while successfully preserving the structure from the contents. Traditional optimization-based method~\cite{gatys2016image} shows poor synthesis quality.
Learning-based methods~\cite{li2017universal,yoo2019photorealistic} that use fixed decoder parameters at test-time. Since both kinds of methods do not consider translation prior, they show limitations in preserving fine details and make artifacts.

Unlike these, our approach has shown high generalization ability to any unseen input images. On the other hand, while Swapping Autoencoder~\cite{park2020swapping} was trained with a large-scale dataset, but limited to generating plausible results, our results show competitive, even better, results. Our success in fine details of both style and content can be found in~\figref{celebahq_result}, where our method produces the most visually appealing images with more vivid details. For example, the top right shows the closest skin color from style and the exact same texture of hair from content as well as overall clarity outperform the state-of-the-art methods.

\paragraph{Quantitative Evaluation.}
We further evaluate our method with the quantitative results as in~\tabref{quantitative_dtp} on photorealistic style transfer examples, CelebA-HQ dataset, and Flicker-Faces dataset (FFHQ) with metrics of Pieapp~\cite{prashnani2018pieapp} (PE) and Structural Similarity (SSIM)~\cite{wang2004image}. Pieapp is a reference-based quality assessment used for semantic consistency. SSIM index is an error measurement which is computed between the original content images and stylized images. Here, we do not measure WCT~\cite{li2017universal} and WCT2~\cite{yoo2019photorealistic} on CelebA-HQ and FFHQ since WCT requires pre-training on each task. Similarly, since CoCosNet~\cite{zhang2020cross} has been trained on example-based image-to-image translation task, we do not evaluate it on photorealistic style transfer examples. We also show the effect of changing weight $\lambda_\mathrm{c}$ of content and style loss.
In the results, our model significantly outperforms most of the  methods on photorealistic examples under the both evaluation metrics. In particular, our results with $\lambda_\mathrm{c}=1/5$ tend to show better quantitative results, since smaller $\lambda_\mathrm{c}$ produces results with more similar structure to content image, which we analyze in more detail in the supplementary material. Also, results on CelebA-HQ are very close to best metrics. The results indicate significant performance gains with our method in all metrics. Interestingly, though our method is designed for style transfer, our architecture also works on image-to-image translation task. 

\subsection{Ablation Study}
In order to validate the effectiveness of each component in our method, we conduct a comprehensive ablation study. In particular, we analyze the effectiveness of warped feature and image, cycle consistency loss $\mathcal{L}_\mathrm{cyc}$, feature moving average, and generation module in~\figref{ablation_mov}. To validate the effect of the warped feature and image in our model, we conduct ablation experiments by replacing the warped feature with random Gaussian noise and eliminating the residual connection with warped image. Without the warped feature, it fails to preserve edges while the result without the warped image fails to capture any style information. 
We also validate the influence of cycle consistency loss, which makes the optimization process more stable. Feature moving average makes the synthesized image more vivid because the previous synthesized feature can give guidance for content and style feature correlation. Without the generation module, the output cannot converge to the optimal result. With all these components, our work is more effective in resulting style relevant outputs while preserving content clearly.
\begin{figure}[t]
	\centering
	\renewcommand{\thesubfigure}{}
	\subfigure[]
	{\includegraphics[width=1\linewidth]{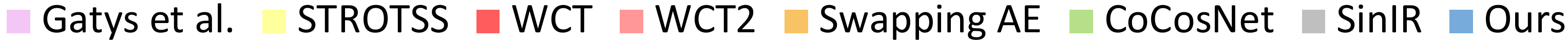}}\hfill\\
	\vspace{-15pt}
	\subfigure[(a)]
	{\includegraphics[width=0.333\linewidth]{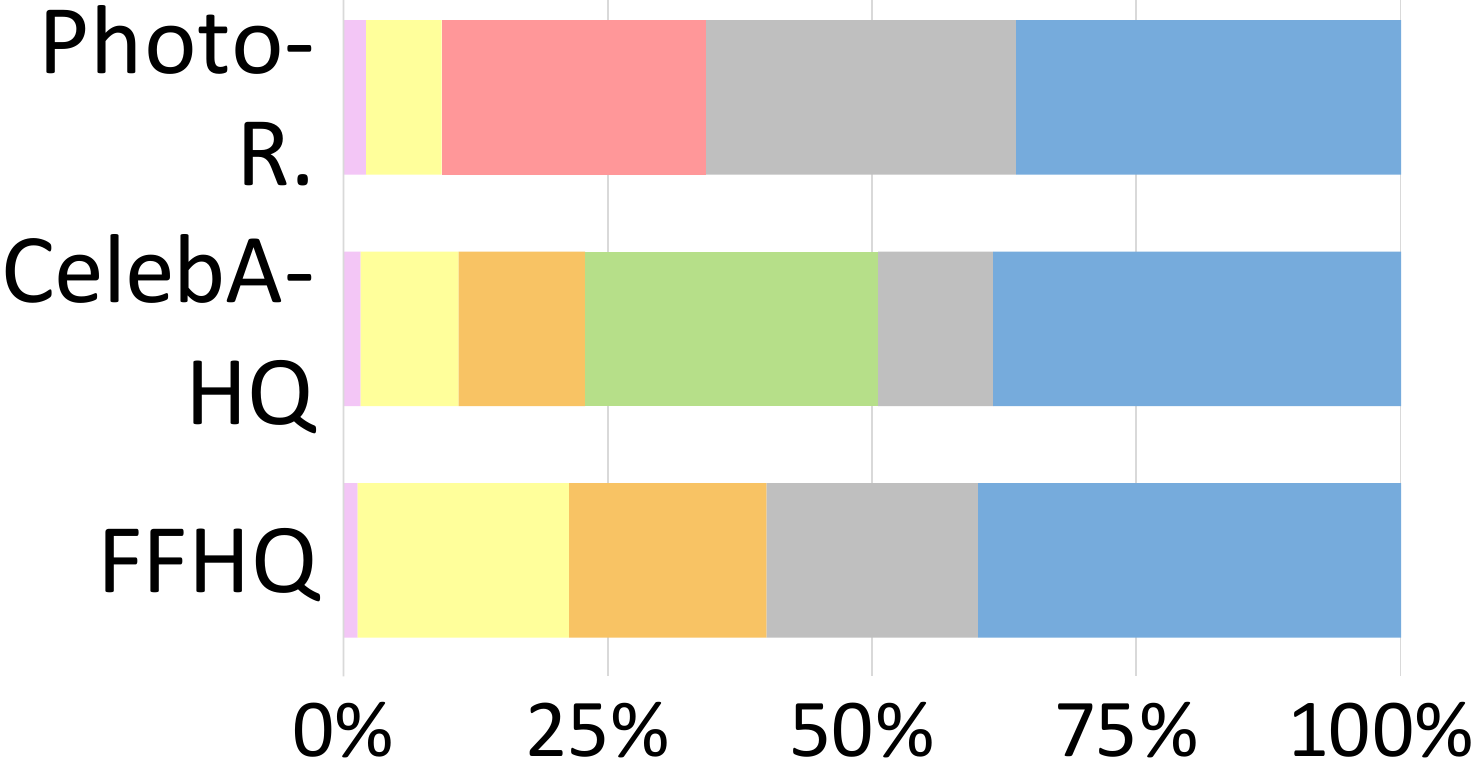}}\hfill
	\subfigure[(b)]
	{\includegraphics[width=0.333\linewidth]{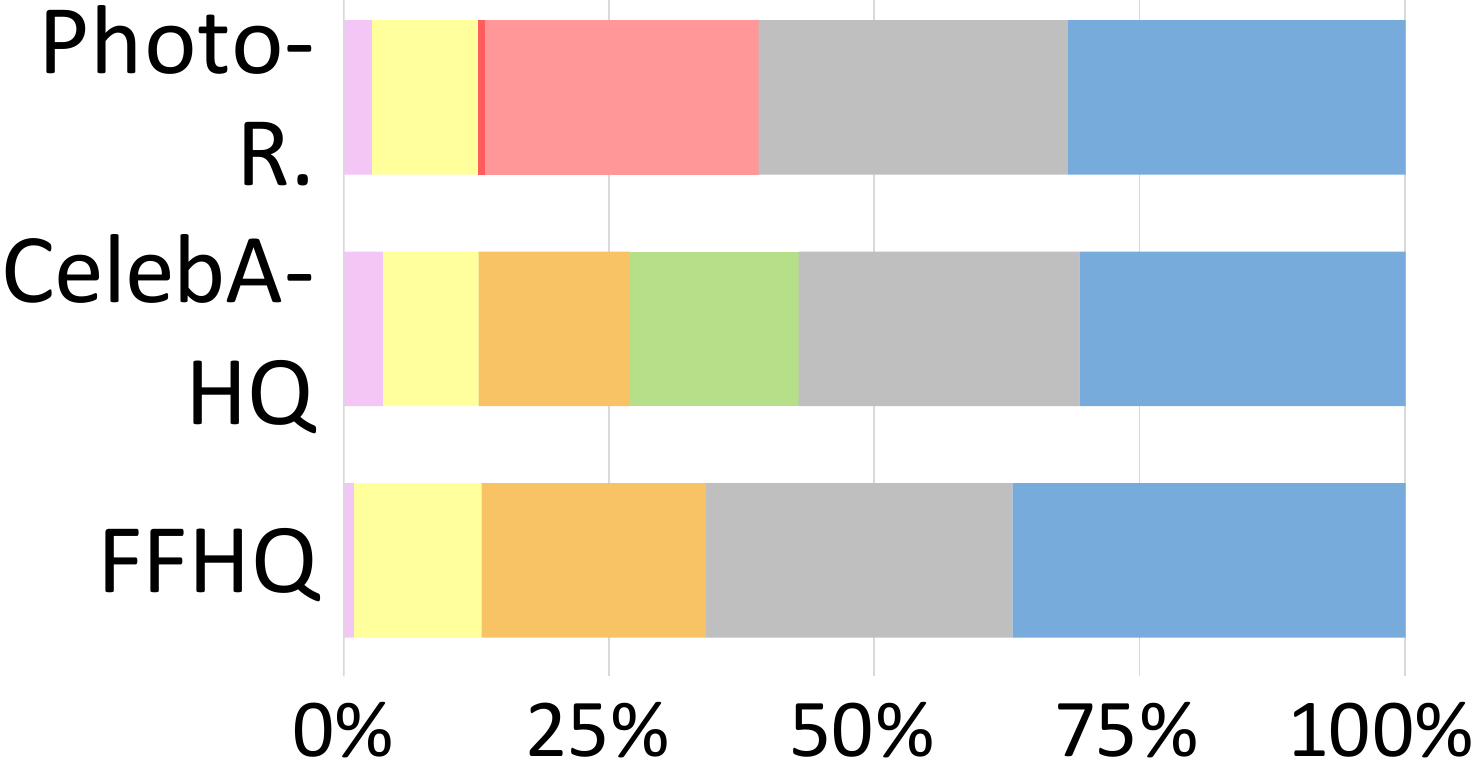}}\hfill
	\subfigure[(c)]
	{\includegraphics[width=0.333\linewidth]{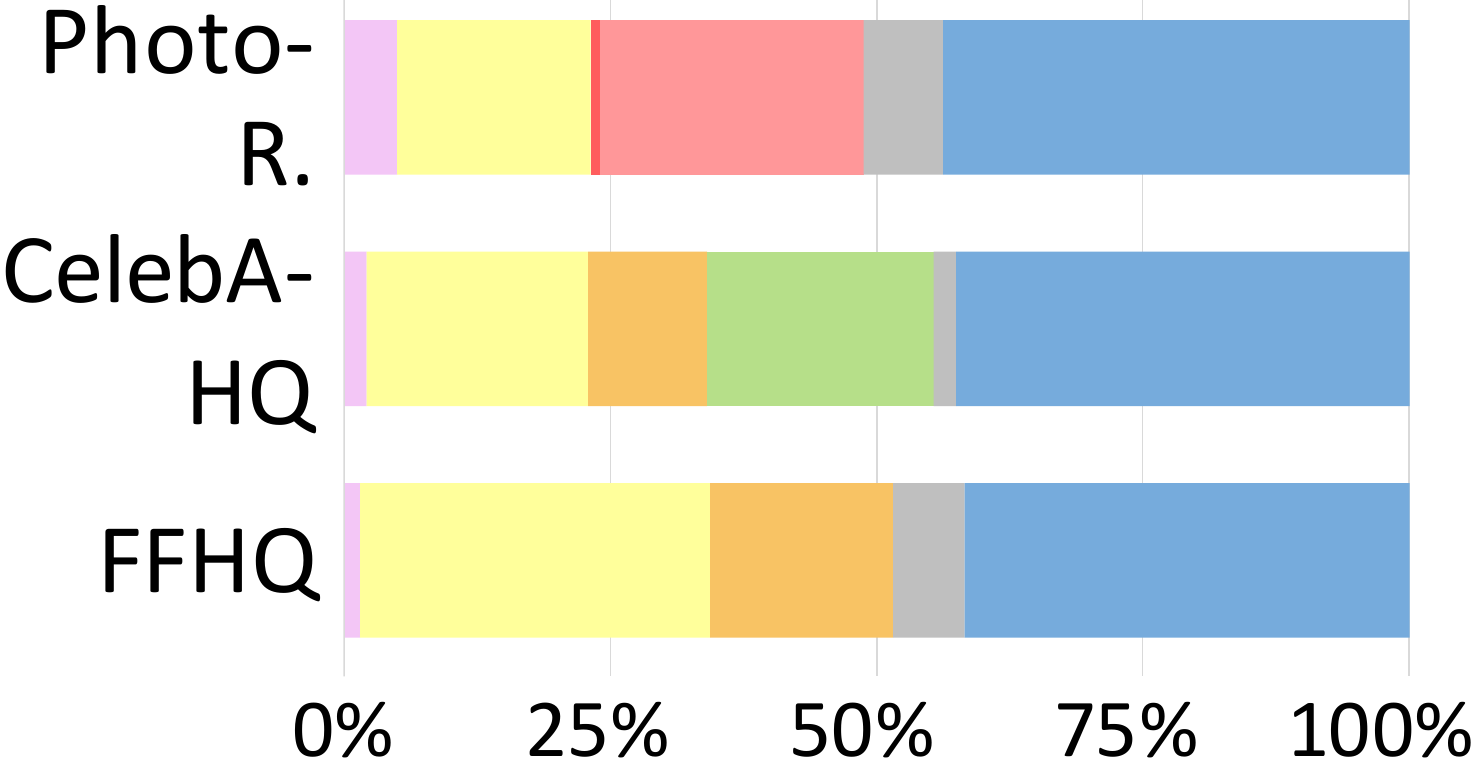}}\hfill\\
	\vspace{-10pt}
	\caption{\textbf{User study results:} (a) image quality, (b) content relevance, and (c) style relevance.}\label{fig_userstudy}\vspace{-10pt}   
\end{figure} 

\begin{comment}
\begin{figure}[t]
	\centering
	\renewcommand{\thesubfigure}{}
	\subfigure[]
	{\includegraphics[width=1\linewidth]{figure/fig8/figure_sh_userstudy_0.pdf}}\hfill\\
	\vspace{-15pt}
	\subfigure[(a)]
	{\includegraphics[width=0.333\linewidth]{figure/fig8/figure_sh_userstudy_1.pdf}}\hfill
	\subfigure[(b)]
	{\includegraphics[width=0.333\linewidth]{figure/fig8/figure_sh_userstudy_2.pdf}}\hfill
	\subfigure[(c)]
	{\includegraphics[width=0.333\linewidth]{figure/fig8/figure_sh_userstudy_3.pdf}}\hfill\\
	\vspace{-10pt}
	\caption{\textbf{User study results for ablation study:} (a) image quality, (b) content relevance, and (c) style relevance.}\label{fig_userstudy_abl}\vspace{-10pt}   
\end{figure} 
\end{comment}

\subsection{User Study}
We also conducted a user study on 80 participants to evaluate the quality of synthesized images in the experiments with the following questions: \textit{``Which do you think has better image quality / similar content to content image / style relavance to style image?''}. On photorealistic style transfer examples, CelebA-HQ~\cite{liu2015deep} and FFHQ dataset~\cite{karras2019style}, our method ranks the first in every cases, which can be found in~\figref{fig_userstudy}. 

\section{Conclusion}
In this paper, we proposed, for the first time, a novel framework to learn the style transfer network on a given input image pair at test-time, without need of any large-scale dataset, hand-labeling, and task-specific training process, called Deep Translation Prior (DTP). Tailored for such test-time training for style transfer, we formulate overall networks as two sub-modules, including correspondence module and generation module.
By training the untrained networks with explicit loss functions for style transfer at test-time, our approach achieves better generalization ability to unseen image pairs or style, which has been one of the major bottlenecks of previous methods. Experimental results on a variety of benchmarks and in comparison to state-of-the-art methods proved that our framework outperforms the existing optimization- and learning-based solutions.
%\textcolor[rgb]{0.00,0.50,0.00}{On the other hand, applications of style transfer could be misused for deception. Addressing this problem, GAN image forensics \cite{farid2016photo, wang2019cnngenerated} have suggested to validate whether the distribution of image is similar to the reals or fakes.}

% \newpage
\newpage
\section*{Acknowledgements}
This research was supported by the MSIT, Korea (IITP-2021-2020-0-01819, ICT Creative Consilience program, and IITP-2021-0-00155, Context and Activity Analysis-based Solution for Safe Childcare) supervised by the IITP and National Research Foundation of Korea (NRF-2021R1C1C1006897). 
{\small
\bibliography{egbib}
}

% \newpage
% \onecolumn

\newpage
\clearpage

\onecolumn
\maketitle
\appendix
\section{Supplementary Material}
In this section, we describe detailed network architecture and provide pseudo-code, additional qualitative results and quantitative results.

\subsection{Network architecture of DTP}
%\subsection{Additional visual results}
We first summarize the detailed network architecture of our DTP in Table~\ref{tab:1}.
Specifically, every convolution layer is followed by ReLU activation. The parameters, $(\mathtt{in},\mathtt{out},\mathtt{k},\mathtt{s},\mathtt{p})$, are the number of input channels, number of output channels, size of the kernel, stride and padding, respectively, in the convolution operation. 
%To utilize richer information when computing global correlation matrix between source and target, we up-sample deep features and concatenate it to corresponding ones. 
`Upsample' indicates bilinearly upsampling the features from the previous layer. 

% \cite{gu2020image}.

\begin{table}[h] %표 수정 필요
\centering
 \scalebox{1.0}{
	
		\begin{tabular}{>{\centering}m{0.18\linewidth} >{\centering}m{0.30\linewidth} 
				>{\centering}m{0.30\linewidth}}
			\hlinewd{0.8pt}
			Layer & Parameters $(\mathtt{in},\mathtt{out},\mathtt{k},\mathtt{s},\mathtt{p})$ & Output shape $(C \times H \times W)$ \tabularnewline
			\hline
			Conv-1-1 &$(3,64,3,1,1)$ & $(64,256,256)$  \tabularnewline
			Conv-1-2 &$(64,64,3,1,1)$ & $(64,256,256)$  \tabularnewline
			MaxPool-1 & - & $(64,128,128)$  \tabularnewline		\hline
			Conv-2-1 & $(64,128,3,1,1)$ & $(128,128,128)$ \tabularnewline
			Conv-2-2 & $(128,128,3,1,1)$ & $(128,128,128)$ \tabularnewline
			MaxPool-2 & - & $(128,64,64)$ \tabularnewline \hline
			Conv-3-1 & $(128,256,3,1,1)$ & $(256,64,64)$ \tabularnewline
			Conv-3-2 & $(256,256,3,1,1)$ & $(256,64,64)$ \tabularnewline
			Conv-3-3 & $(256,256,3,1,1)$ & $(256,64,64)$ \tabularnewline
			Conv-3-4 & $(256,256,3,1,1)$ & $(256,64,64)$ \tabularnewline
			MaxPool-3 & - & $(256,32,32)$ \tabularnewline \hline
			Conv-4-1 & $(256,512,3,1,1)$ & $(512,32,32)$ \tabularnewline
			Conv-4-2 & $(512,512,3,1,1)$ & $(512,32,32)$ \tabularnewline
			Conv-4-3 & $(512,512,3,1,1)$ & $(512,32,32)$ \tabularnewline
			Conv-4-4 & $(512,512,3,1,1)$ & $(512,32,32)$ \tabularnewline
			MaxPool-4 & - & $(512,16,16)$ \tabularnewline \hline
			Conv-5-1 & $(512,512,3,1,1)$ & $(512,16,16)$ \tabularnewline
			Conv-5-2 & $(512,512,3,1,1)$ & $(512,16,16)$ \tabularnewline
            \hline
            % Up (Conv-3-4) & $(256,256,3,1,1)$ & $(256,64,64)$ \tabularnewline
            %Concat1 (Conv-4-2) & - & $(1024,32,32)$ \tabularnewline
            Up (Conv-4-3) & $(256,256,3,1,1)$ & $(256,64,64)$ \tabularnewline
            %Concat2 (Conv-3-2) & - & $(1280,64,64)$ \tabularnewline
            Up (Conv-4-2) & $(256,256,3,1,1)$ & $(256,64,64)$ \tabularnewline
            Up (Conv-4-1) & $(256,256,3,1,1)$ & $(128,64,64)$ \tabularnewline 
            Upsample & - & $(128,128,128)$ \tabularnewline \hline
            Up (Conv-3-4) & $(128,128,3,1,1)$ & $(128,128,128)$ \tabularnewline
            Up (Conv-3-3) & $(128,128,3,1,1)$ & $(128,128,128)$ \tabularnewline
            Up (Conv-3-2) & $(128,128,3,1,1)$& $(128,128,128)$ \tabularnewline
            Up (Conv-3-1) & $(128,64,3,1,1)$& $(64,128,128)$ \tabularnewline 
            Upsample & - & $(64,256,256)$ \tabularnewline \hline
            Up (Conv-2-3) & $(64,64,3,1,1)$ & $(64,256,256)$ \tabularnewline
            Up (Conv-2-2) & $(64,64,3,1,1)$ & $(64,256,256)$ \tabularnewline
            Up (Conv-2-1) & $(64,32,3,1,1)$ & $(32,256,256)$ \tabularnewline
            \hline
            Up (Conv-1-3) & $(32,32,3,1,1)$ & $(32,256,256)$ \tabularnewline
            Up (Conv-1-2) & $(32,32,3,1,1)$ & $(3,256,256)$ \tabularnewline
            Up (Conv-1-1) & $(32,3,3,1,1)$ & $(3,256,256)$ \tabularnewline

            \hline
            Correlation & - & $(1,4096,4096)$ \tabularnewline
            Warping & - & $(3,256,256)$ \tabularnewline

			\hlinewd{0.8pt}
		\end{tabular}
	}
	\caption{\textbf{Network architecture of our DTP.}}
	\label{tab:1}
\end{table}

\newpage
\subsection{Pseudo-code.} 
\begin{figure*}[h]
    \centering
    \begin{center}
	{\includegraphics[width=0.7\linewidth]{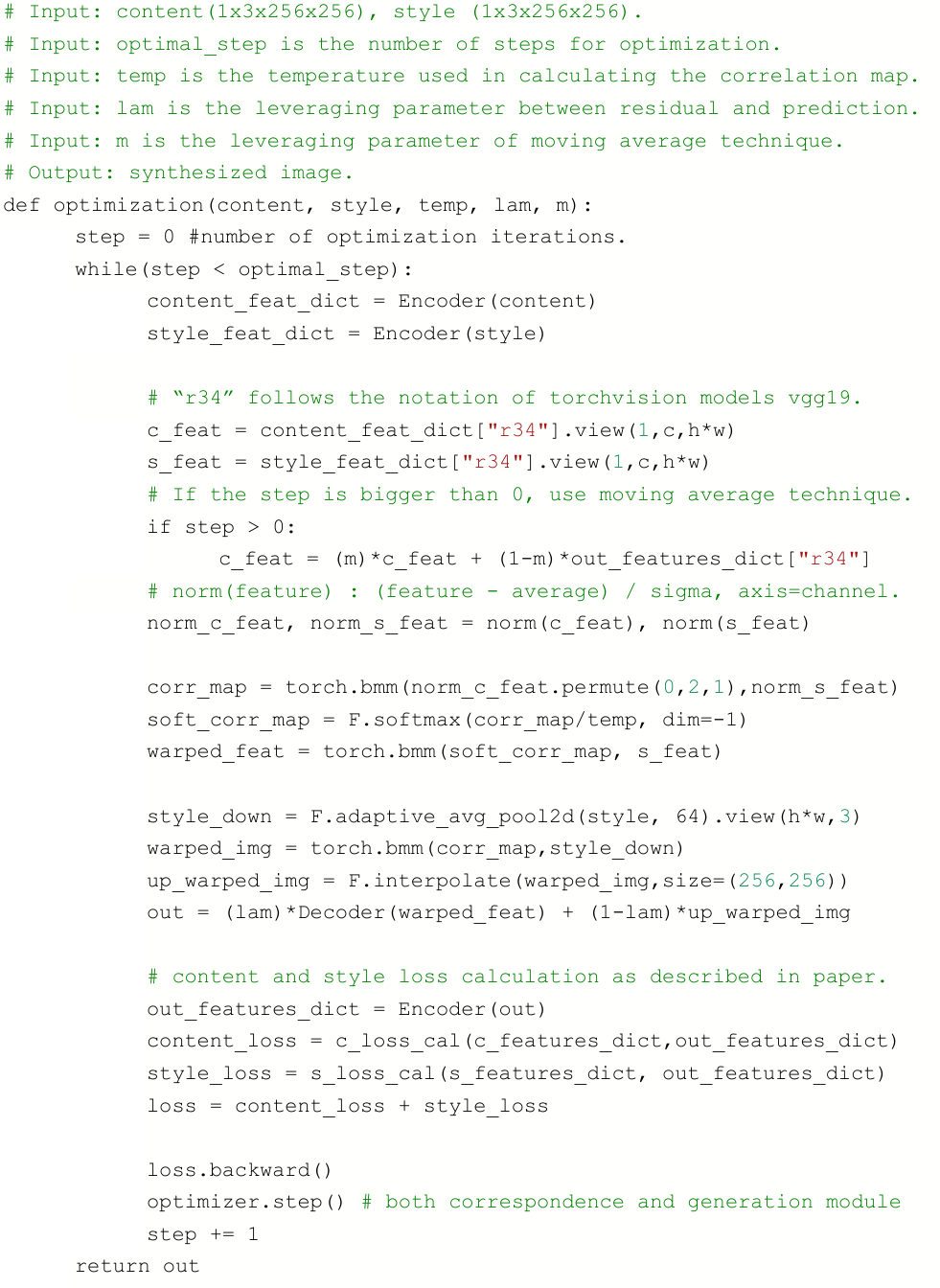}}\hfill
	\end{center}
   %\vspace{-5pt}
	\caption{\textbf{PyTorch-like pseudo-code.}}
	\label{fig:code}\vspace{-10pt}

\end{figure*}

Here we provide the (PyTorch-like) pseudo-code for DTP in \figref{fig:code}. 
%\inputminted[fontsize=\scriptsize]{python}{figure/dtp.py}
\vspace{5pt}

\newpage

\subsection{User study on Ablation Study.} 
We conducted user study on ablation study, as shown in~\tabref{table:ablation_user_study}. Total 40 sets of images were asked to 52 subjects, in which subjects had to choose one among six synthesized images.
\begin{table}[h]
\centering
\scalebox{0.8}{
\begin{tabular}{l|cccccc}
\hline
                 & -WF    & -WI   & -FMA   & -$\mathcal{L}_\mathrm{cyc}$  &  -G. Module. & Ours   \\ \hline
Content relevance & 0.0\% & 15.2\% & 4.0\% & 21.1\% & 0.7\%  & \textbf{59.0\%} \\
Best stylization & 17.4\% & 1.3\% & 2.9\% & 0.5\% & 1.0\%  & \textbf{76.9\%} \\
Most preferred   & 1.8\% & 3.2\% & 1.5\% & 3.6\% & 0.1\%  & \textbf{89.8\%} \\ \hline
\end{tabular}}\vspace{-5pt}
\caption{{User study results on ablation study.} 
}
\label{table:ablation_user_study}\vspace{-10pt}
\end{table}

\subsection{Additional Visual Results.} 
\begin{figure*}[h]
	\begin{center}
	\renewcommand{\thesubfigure}{}
	\subfigure[]
	{\includegraphics[width=0.163\linewidth]{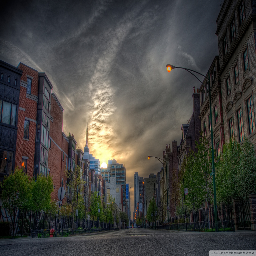}}\hfill
    \subfigure[]
	{\includegraphics[width=0.163\linewidth]{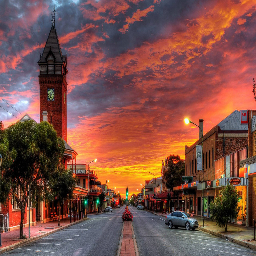}}\hfill
	\subfigure[]	
	{\includegraphics[width=0.163\linewidth]{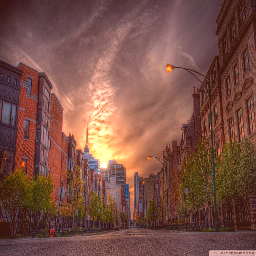}}
	\subfigure[]	
	{\includegraphics[width=0.163\linewidth]{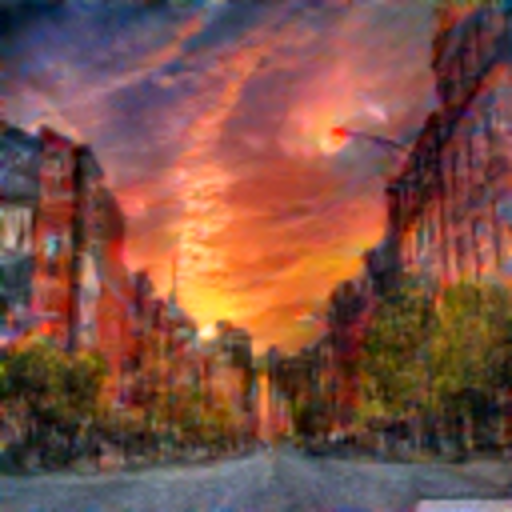}}
	\subfigure[]	
	{\includegraphics[width=0.163\linewidth]{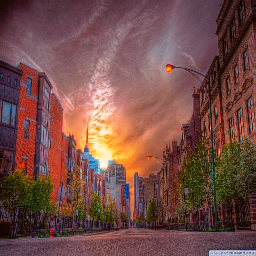}}
	\subfigure[]	
	{\includegraphics[width=0.163\linewidth]{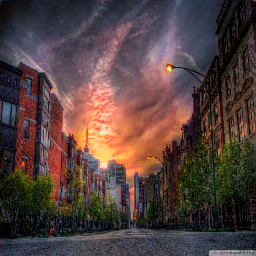}}\hfill\\\vspace{-20.5pt}
	
	\subfigure[]
	{\includegraphics[width=0.163\linewidth]{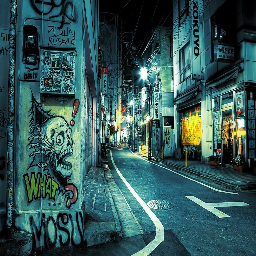}}\hfill
    \subfigure[]
	{\includegraphics[width=0.163\linewidth]{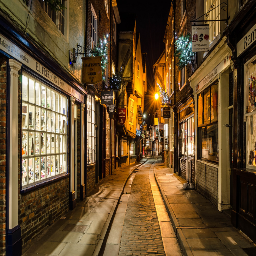}}\hfill
	\subfigure[]
	{\includegraphics[width=0.163\linewidth]{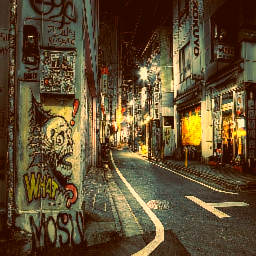}}\hfill
    \subfigure[]
	{\includegraphics[width=0.163\linewidth]{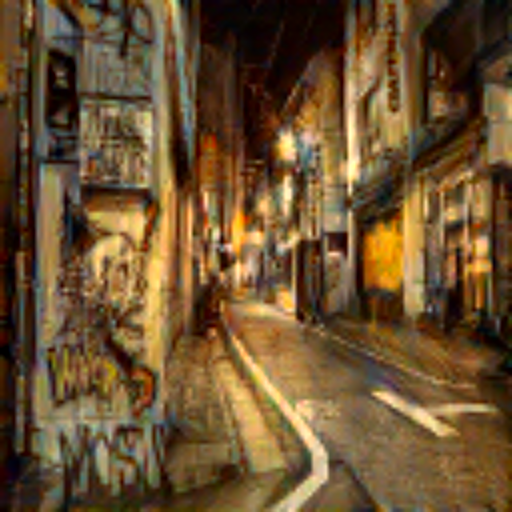}}\hfill
	\subfigure[]
	{\includegraphics[width=0.163\linewidth]{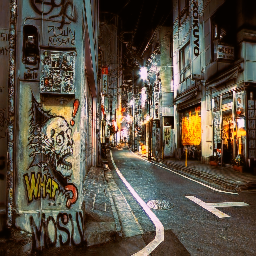}}\hfill
	\subfigure[]	
	{\includegraphics[width=0.163\linewidth]{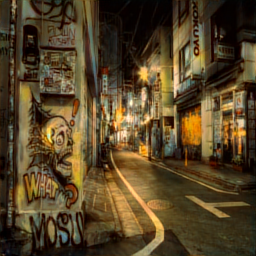}}\hfill\\\vspace{-20.5pt}
	
	\subfigure[]
	{\includegraphics[width=0.163\linewidth]{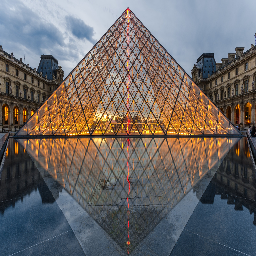}}\hfill
    \subfigure[]
	{\includegraphics[width=0.163\linewidth]{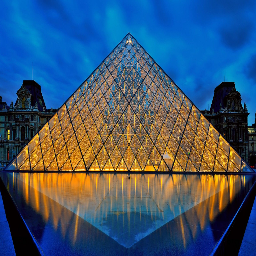}}\hfill
	\subfigure[]
	{\includegraphics[width=0.163\linewidth]{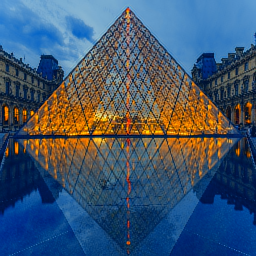}}\hfill
    \subfigure[]
	{\includegraphics[width=0.163\linewidth]{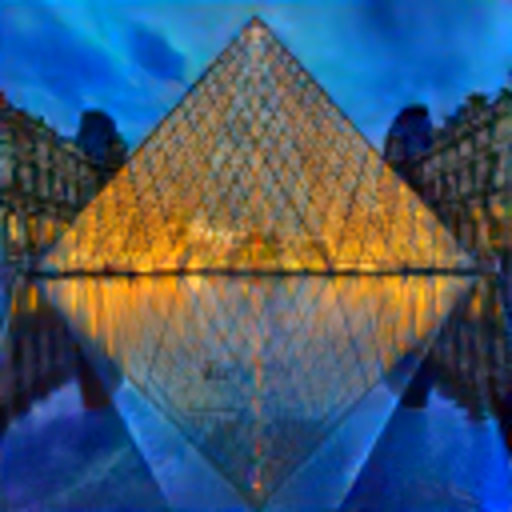}}\hfill
	\subfigure[]
	{\includegraphics[width=0.163\linewidth]{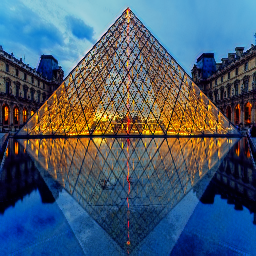}}\hfill
	\subfigure[]	
	{\includegraphics[width=0.163\linewidth]{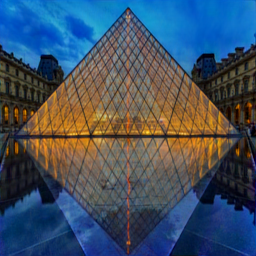}}\hfill\\\vspace{-20.5pt}

% 	\subfigure[]
% 	{\includegraphics[width=0.163\linewidth]{figure/supp/fig10/14/in14.png}}\hfill
%     \subfigure[]
% 	{\includegraphics[width=0.163\linewidth]{figure/supp/fig10/14/tar14.png}}\hfill
% 	\subfigure[]	
% 	{\includegraphics[width=0.163\linewidth]{figure/supp/fig10/14/14.png}}\hfill
% 	\subfigure[]	
% 	{\includegraphics[width=0.163\linewidth]{figure/supp/fig10/14/strotss14.png}}\hfill
% 	\subfigure[]	
% 	{\includegraphics[width=0.163\linewidth]{figure/supp/fig10/14/sinir14.png}}\hfill
% 	\subfigure[]	
% 	{\includegraphics[width=0.163\linewidth]{figure/supp/fig10/14/result.png}}\hfill\\\vspace{-20.5pt}

	\subfigure[]
	{\includegraphics[width=0.163\linewidth]{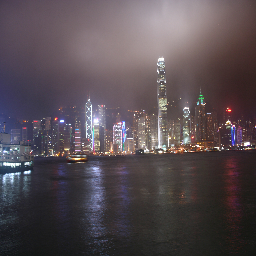}}\hfill
    \subfigure[]
	{\includegraphics[width=0.163\linewidth]{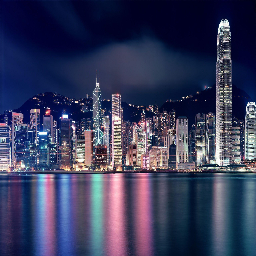}}\hfill
	\subfigure[]	
	{\includegraphics[width=0.163\linewidth]{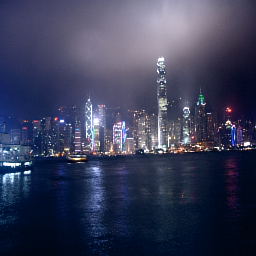}}\hfill
	\subfigure[]	
	{\includegraphics[width=0.163\linewidth]{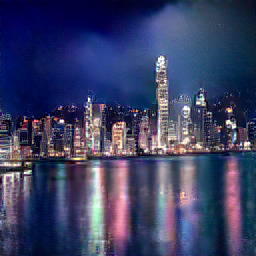}}\hfill
	\subfigure[]	
	{\includegraphics[width=0.163\linewidth]{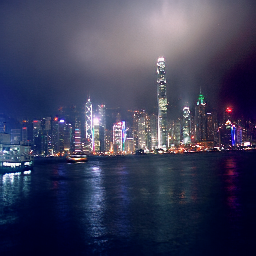}}\hfill
	\subfigure[]	
	{\includegraphics[width=0.163\linewidth]{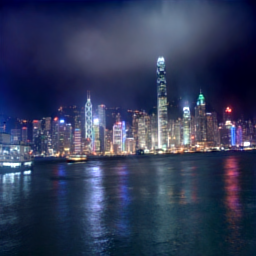}}\hfill\\
	\vspace{-20.5pt}
	\subfigure[Content]
	{\includegraphics[width=0.163\linewidth]{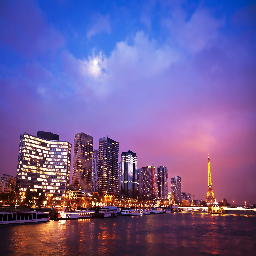}}\hfill
    \subfigure[Style]
	{\includegraphics[width=0.163\linewidth]{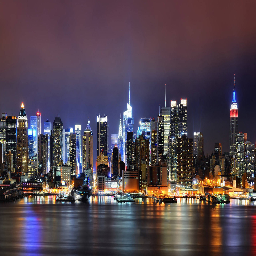}}\hfill
	\subfigure[WCT2]
	{\includegraphics[width=0.163\linewidth]{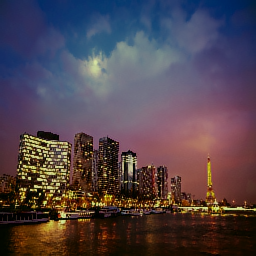}}\hfill
    \subfigure[STROTSS]
	{\includegraphics[width=0.163\linewidth]{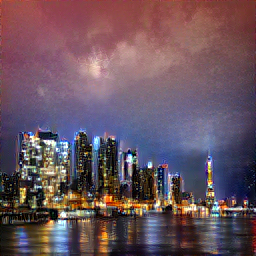}}\hfill
	\subfigure[SinIR]
	{\includegraphics[width=0.163\linewidth]{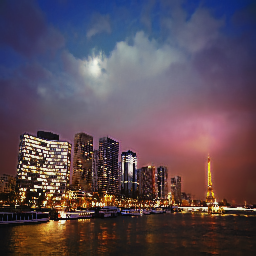}}\hfill
	\subfigure[DTP Results]	
	{\includegraphics[width=0.163\linewidth]{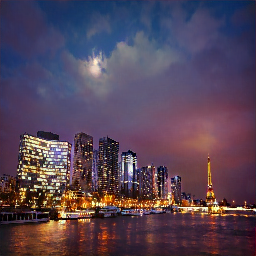}}\hfill\\

    \vspace{-5pt}
	\caption{\textbf{Results on standard photorealistic style transfer benchmarks~\cite{an2019ultrafast}. Given an input pair, we compare the results with WCT2~\cite{yoo2019photorealistic}, STROTSS~\cite{kolkin2019style},  SinIR~\cite{yoo2021sinir} and ours (DTP). }}
	\label{fig:stylenas}\vspace{-10pt}
	    \end{center}
\end{figure*}

\begin{figure*}[h]
	\begin{center}
	\renewcommand{\thesubfigure}{}
	\subfigure[]
	{\includegraphics[width=0.163\linewidth]{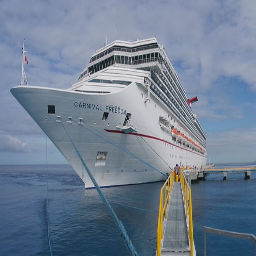}}\hfill
    \subfigure[]
	{\includegraphics[width=0.163\linewidth]{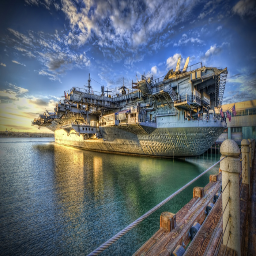}}\hfill
	\subfigure[]
	{\includegraphics[width=0.163\linewidth]{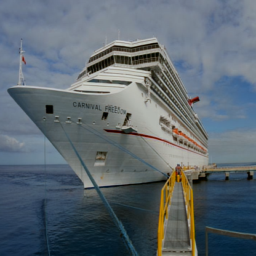}}\hfill
    \subfigure[]
	{\includegraphics[width=0.163\linewidth]{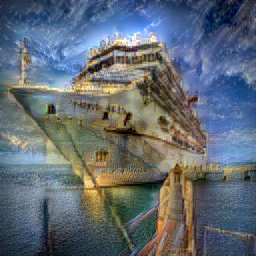}}\hfill
    \subfigure[]
	{\includegraphics[width=0.163\linewidth]{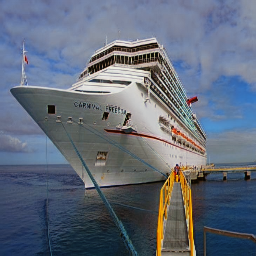}}\hfill
	\subfigure[]	
	{\includegraphics[width=0.163\linewidth]{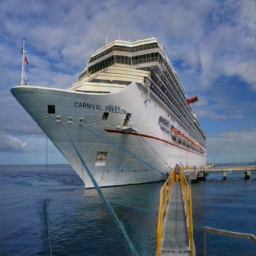}}\hfill\\%
	\vspace{-20.5pt}
	\subfigure[]
	{\includegraphics[width=0.163\linewidth]{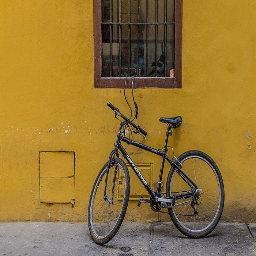}}\hfill
    \subfigure[]
	{\includegraphics[width=0.163\linewidth]{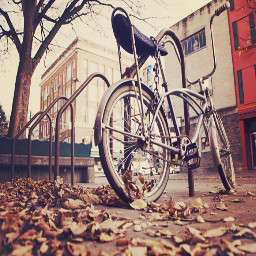}}\hfill
	\subfigure[]
	{\includegraphics[width=0.163\linewidth]{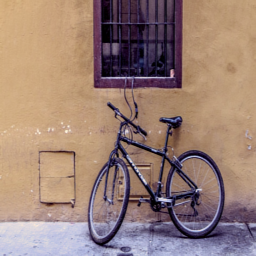}}\hfill
    \subfigure[]
	{\includegraphics[width=0.163\linewidth]{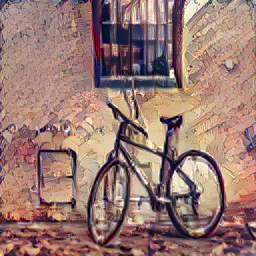}}\hfill
    \subfigure[]
	{\includegraphics[width=0.163\linewidth]{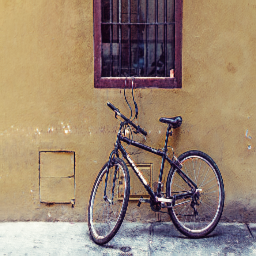}}\hfill
	\subfigure[]	
	{\includegraphics[width=0.163\linewidth]{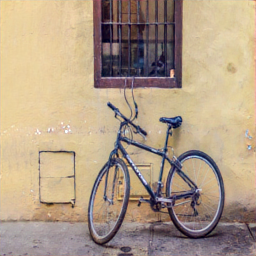}}\hfill\\%
	\vspace{-20.5pt}
	\subfigure[]
	{\includegraphics[width=0.163\linewidth]{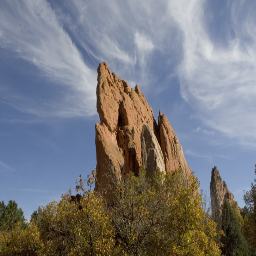}}\hfill
    \subfigure[]
	{\includegraphics[width=0.163\linewidth]{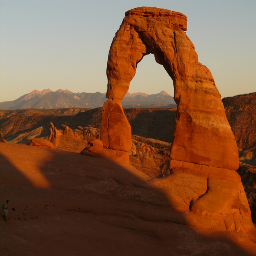}}\hfill
	\subfigure[]
	{\includegraphics[width=0.163\linewidth]{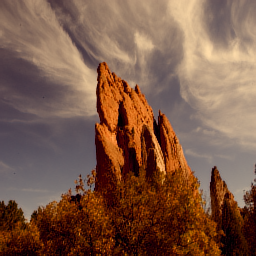}}\hfill
    \subfigure[]
	{\includegraphics[width=0.163\linewidth]{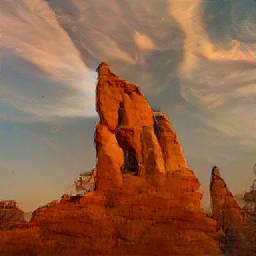}}\hfill
    \subfigure[]
	{\includegraphics[width=0.163\linewidth]{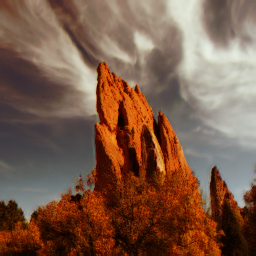}}\hfill
	\subfigure[]	
	{\includegraphics[width=0.163\linewidth]{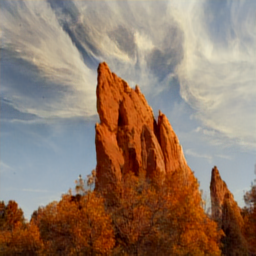}}\hfill\\%
	\vspace{-20.5pt}
	\subfigure[]
	{\includegraphics[width=0.163\linewidth]{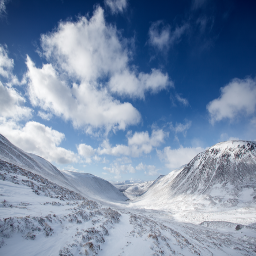}}\hfill
    \subfigure[]
	{\includegraphics[width=0.163\linewidth]{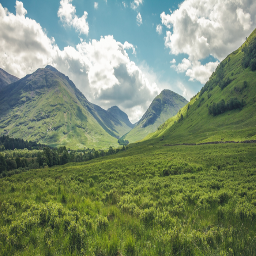}}\hfill
	\subfigure[]
	{\includegraphics[width=0.163\linewidth]{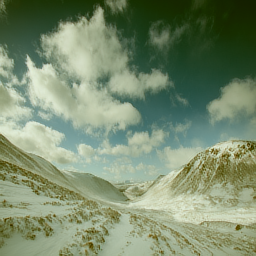}}\hfill
    \subfigure[]
	{\includegraphics[width=0.163\linewidth]{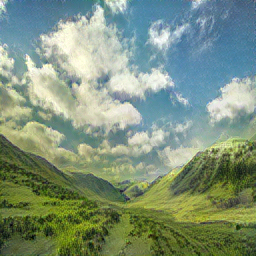}}\hfill
    \subfigure[]
	{\includegraphics[width=0.163\linewidth]{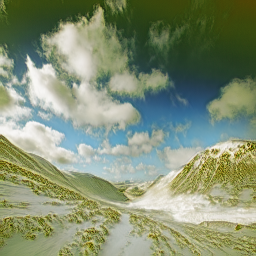}}\hfill
	\subfigure[]	
	{\includegraphics[width=0.163\linewidth]{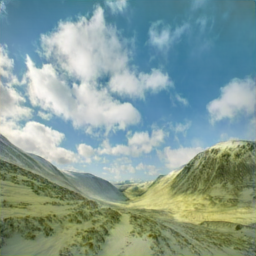}}\hfill\\%
	\vspace{-20.5pt}
	\subfigure[]
	{\includegraphics[width=0.163\linewidth]{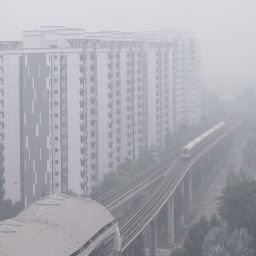}}\hfill
	\subfigure[]
	{\includegraphics[width=0.163\linewidth]{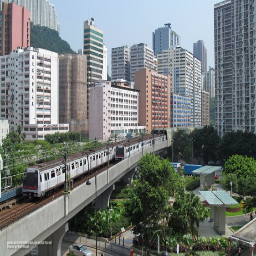}}\hfill
	\subfigure[]
	{\includegraphics[width=0.163\linewidth]{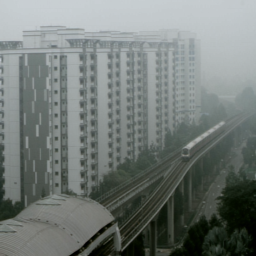}}\hfill
	\subfigure[]
	{\includegraphics[width=0.163\linewidth]{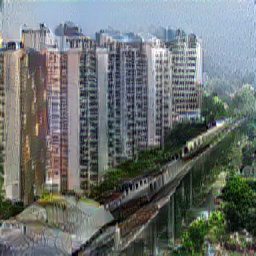}}\hfill
	\subfigure[]
	{\includegraphics[width=0.163\linewidth]{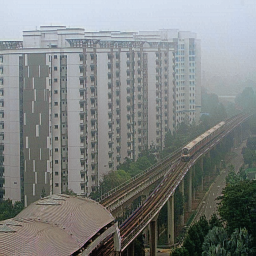}}\hfill
	\subfigure[]
	{\includegraphics[width=0.163\linewidth]{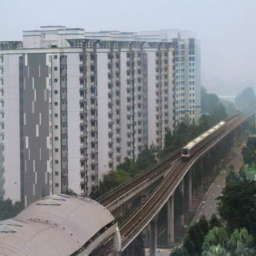}}\hfill\\\vspace{-20.5pt}
	
	\subfigure[Content]%
	{\includegraphics[width=0.163\linewidth]{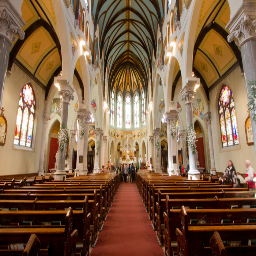}}\hfill
    \subfigure[Style]
	{\includegraphics[width=0.163\linewidth]{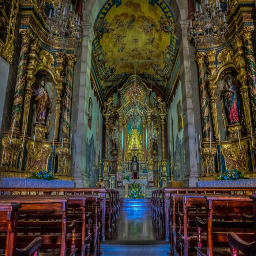}}\hfill
	\subfigure[WCT2]
	{\includegraphics[width=0.163\linewidth]{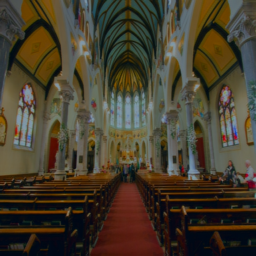}}\hfill
    \subfigure[STROTSS]
	{\includegraphics[width=0.163\linewidth]{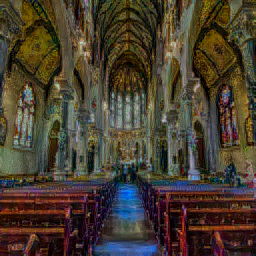}}\hfill
    \subfigure[SinIR]
	{\includegraphics[width=0.163\linewidth]{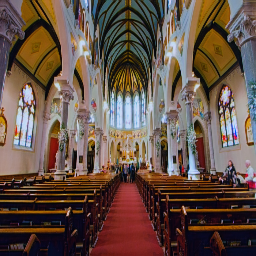}}\hfill
	\subfigure[DTP Results]
	{\includegraphics[width=0.163\linewidth]{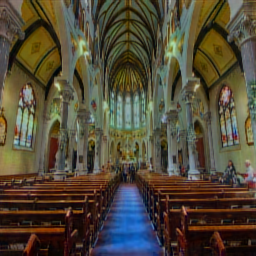}}\hfill\\

    \vspace{-5pt}
	\caption{\textbf{Results on standard photorealistic style transfer benchmarks~\cite{luan2017deep}. Given an input pair, we compare the results with WCT2~\cite{yoo2019photorealistic}, STROTSS~\cite{kolkin2019style},  SinIR~\cite{yoo2021sinir} and ours (DTP).}}
	\label{fig:photo}\vspace{-10pt}
	    \end{center}
\end{figure*}

\begin{figure*}[t]
	\begin{center}
	\renewcommand{\thesubfigure}{}
	\subfigure[]
	{\includegraphics[width=0.163\linewidth]{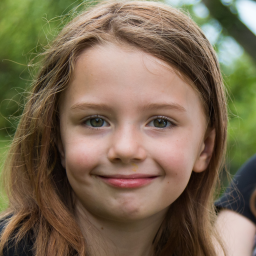}}\hfill
	\subfigure[]
	{\includegraphics[width=0.163\linewidth]{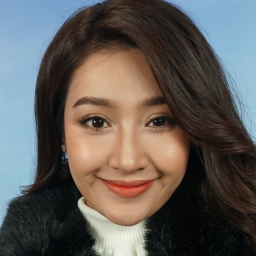}}\hfill
	\subfigure[]
	{\includegraphics[width=0.163\linewidth]{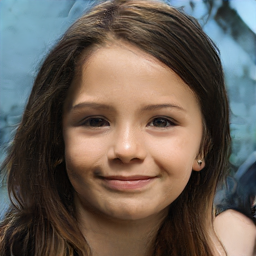}}\hfill
	\subfigure[]
	{\includegraphics[width=0.163\linewidth]{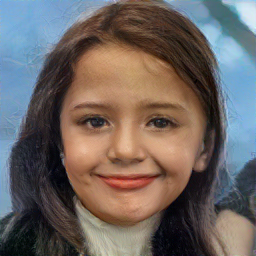}}\hfill
	\subfigure[]
	{\includegraphics[width=0.163\linewidth]{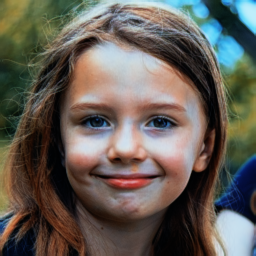}}\hfill
	\subfigure[]
	{\includegraphics[width=0.163\linewidth]{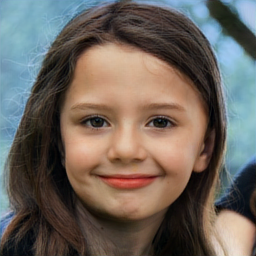}}\hfill\\
	\vspace{-20.5pt}

	\subfigure[]
	{\includegraphics[width=0.163\linewidth]{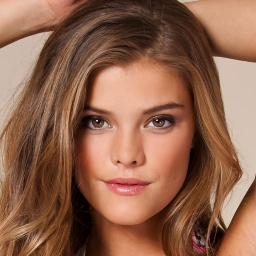}}\hfill
    \subfigure[]
	{\includegraphics[width=0.163\linewidth]{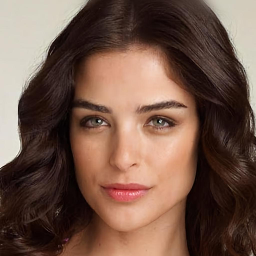}}\hfill
	\subfigure[]
	{\includegraphics[width=0.163\linewidth]{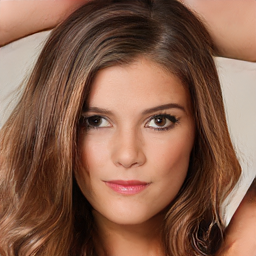}}\hfill
    \subfigure[]
	{\includegraphics[width=0.163\linewidth]{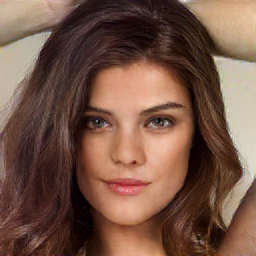}}\hfill
    \subfigure[]
	{\includegraphics[width=0.163\linewidth]{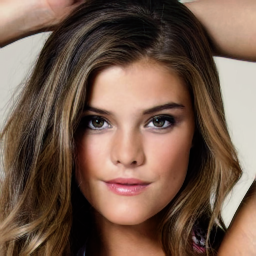}}\hfill
	\subfigure[]	
	{\includegraphics[width=0.163\linewidth]{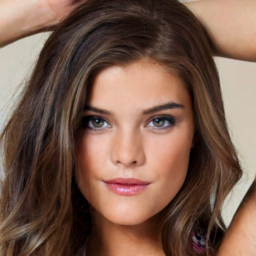}}\hfill\\
	\vspace{-20.5pt}
	
	\subfigure[]
	{\includegraphics[width=0.163\linewidth]{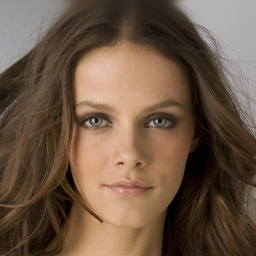}}\hfill
    \subfigure[]
	{\includegraphics[width=0.163\linewidth]{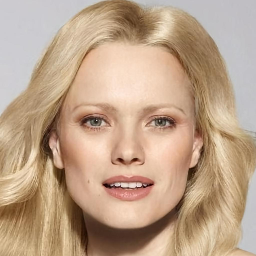}}\hfill
	\subfigure[]
	{\includegraphics[width=0.163\linewidth]{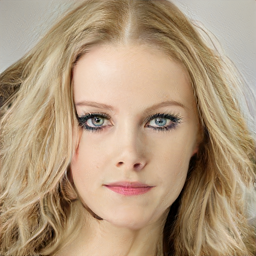}}\hfill
    \subfigure[]
	{\includegraphics[width=0.163\linewidth]{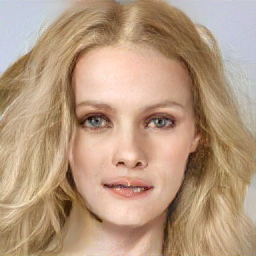}}\hfill
    \subfigure[]
	{\includegraphics[width=0.163\linewidth]{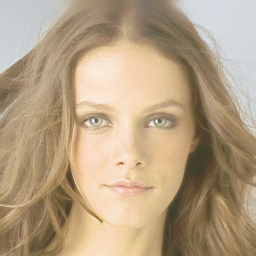}}\hfill
	\subfigure[]	
	{\includegraphics[width=0.163\linewidth]{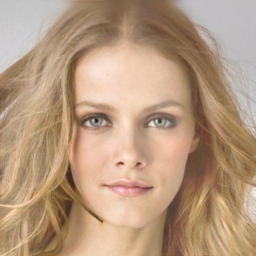}}\hfill\\%
	\vspace{-20.5pt}
	
	\subfigure[]
	{\includegraphics[width=0.163\linewidth]{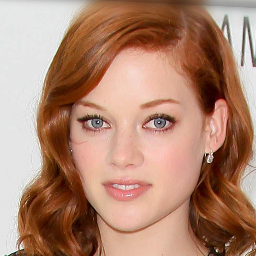}}\hfill
    \subfigure[]
	{\includegraphics[width=0.163\linewidth]{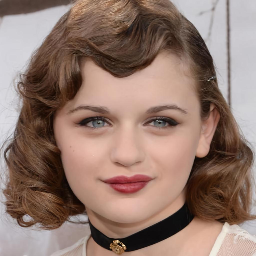}}\hfill
	%\vspace{-5pt}
	\subfigure[]
	{\includegraphics[width=0.163\linewidth]{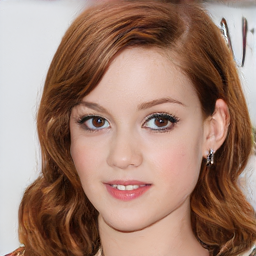}}\hfill
    \subfigure[]
	{\includegraphics[width=0.163\linewidth]{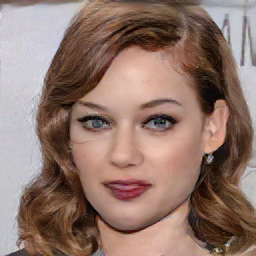}}\hfill
    \subfigure[]
	{\includegraphics[width=0.163\linewidth]{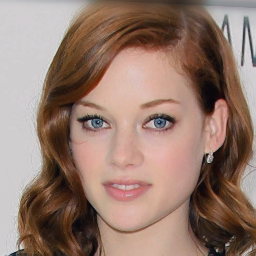}}\hfill
	\subfigure[]
	{\includegraphics[width=0.163\linewidth]{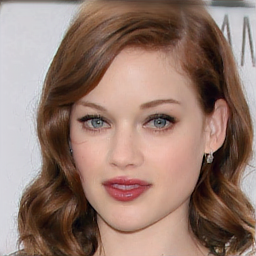}}\hfill\\
	\vspace{-20.5pt}
	\subfigure[]
	{\includegraphics[width=0.163\linewidth]{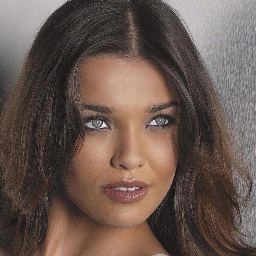}}\hfill
    \subfigure[]
	{\includegraphics[width=0.163\linewidth]{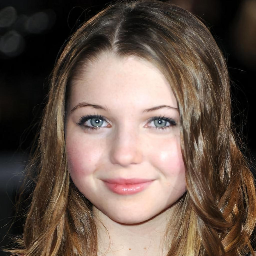}}\hfill
	\subfigure[]
	{\includegraphics[width=0.163\linewidth]{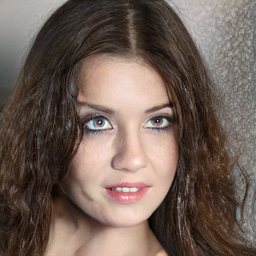}}\hfill
    \subfigure[]
	{\includegraphics[width=0.163\linewidth]{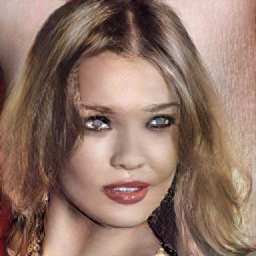}}\hfill
    \subfigure[]
	{\includegraphics[width=0.163\linewidth]{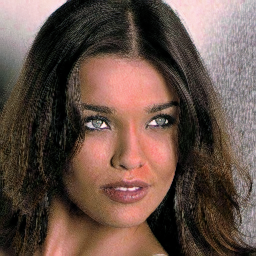}}\hfill
	\subfigure[]	
	{\includegraphics[width=0.163\linewidth]{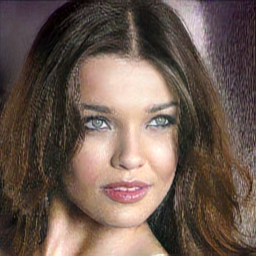}}\hfill\\
	\vspace{-20.5pt}
	\subfigure[Content]
	{\includegraphics[width=0.163\linewidth]{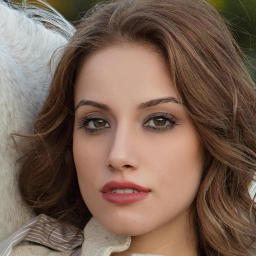}}\hfill
    \subfigure[Style]
	{\includegraphics[width=0.163\linewidth]{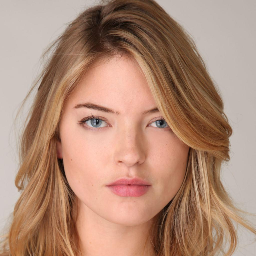}}\hfill
	%\vspace{-5pt}
	\subfigure[Swap AE]
	{\includegraphics[width=0.163\linewidth]{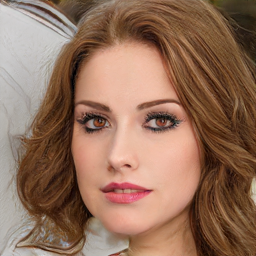}}\hfill
    \subfigure[STROTSS]
	{\includegraphics[width=0.163\linewidth]{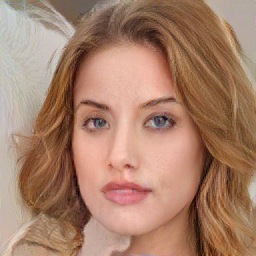}}\hfill
    \subfigure[SinIR]
	{\includegraphics[width=0.163\linewidth]{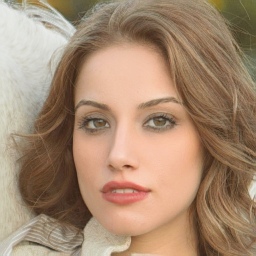}}\hfill
	\subfigure[DTP Results]	
	{\includegraphics[width=0.163\linewidth]{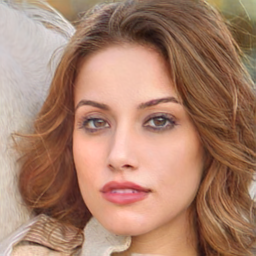}}\hfill\\
    \vspace{-5pt}
	\caption{\textbf{Results on CelebA-HQ ~\cite{liu2015deep} and FFHQ ~\cite{karras2019style} benchmarks. Given an input pair, we compare the results with Swap AE ~\cite{park2020swapping}, STROTSS ~\cite{kolkin2019style}, SinIR ~\cite{yoo2021sinir} and ours (DTP).}}
	\label{fig:celeb}
	\vspace{-10pt}
	    \end{center}
\end{figure*}

In the main paper, we have shown our results of photorealistic style transfer and image-to-image translation. Here, we show additional results in \figref{fig:stylenas}, \figref{fig:photo} and \figref{fig:celeb} to demonstrate the robustness of ours.
%~\figref{celebahq_result}, \figref{ffhq_result}, and \figref{photo_result} of

\subsection{Additional Examples on Ablation Study.} 
\begin{figure*}[t]
\centering
	\renewcommand{\thesubfigure}{}
	\subfigure[]
	{\includegraphics[width=0.122\linewidth]{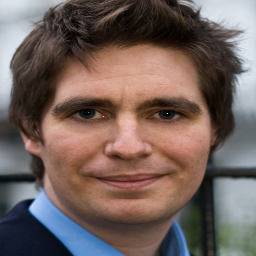}}\hfill
	\subfigure[]
	{\includegraphics[width=0.122\linewidth]{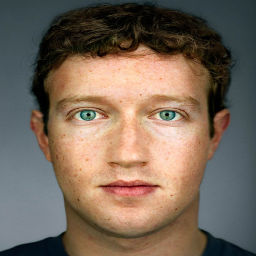}}\hfill
	\subfigure[]
	{\includegraphics[width=0.122\linewidth]{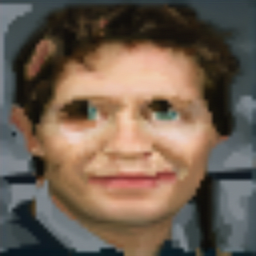}}\hfill
	\subfigure[]
	{\includegraphics[width=0.122\linewidth]{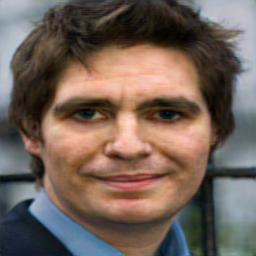}}\hfill
	\subfigure[]
	{\includegraphics[width=0.122\linewidth]{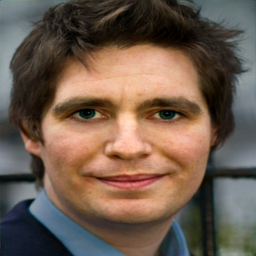}}\hfill
	\subfigure[]
	{\includegraphics[width=0.122\linewidth]{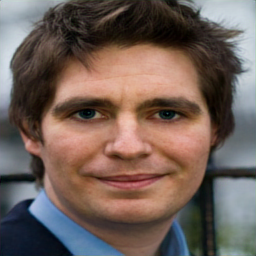}}\hfill
    \subfigure[]
	{\includegraphics[width=0.122\linewidth]{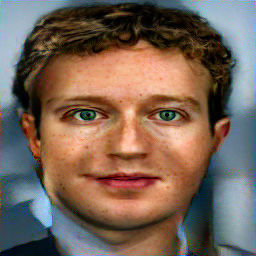}}\hfill
	\subfigure[]
	{\includegraphics[width=0.122\linewidth]{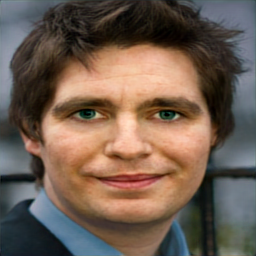}}\hfill\\
	\vspace{-20pt}	

	\subfigure[]
	{\includegraphics[width=0.122\linewidth]{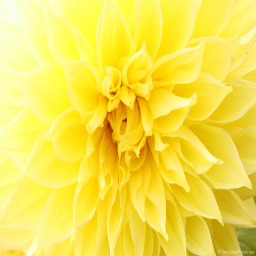}}\hfill
	\subfigure[]
	{\includegraphics[width=0.122\linewidth]{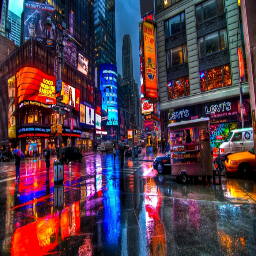}}\hfill
	\subfigure[]
	{\includegraphics[width=0.122\linewidth]{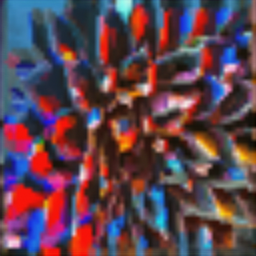}}\hfill
	\subfigure[]
	{\includegraphics[width=0.122\linewidth]{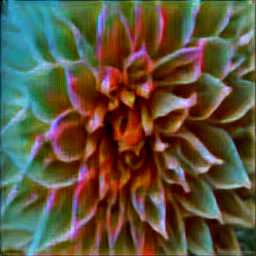}}\hfill
	\subfigure[]
	{\includegraphics[width=0.122\linewidth]{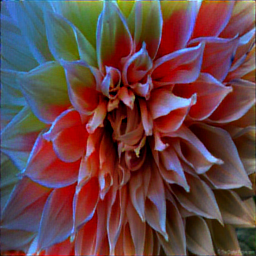}}\hfill
	\subfigure[]
	{\includegraphics[width=0.122\linewidth]{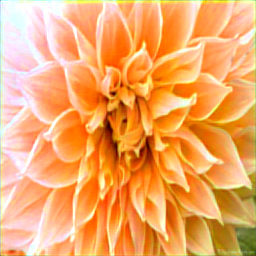}}\hfill
    \subfigure[]
	{\includegraphics[width=0.122\linewidth]{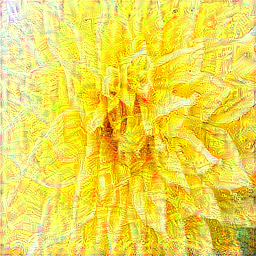}}\hfill
	\subfigure[]
	{\includegraphics[width=0.122\linewidth]{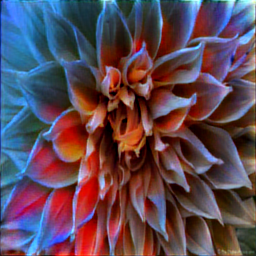}}\hfill\\
	\vspace{-20pt}
	
	\subfigure[]
	{\includegraphics[width=0.122\linewidth]{figure/supp/fig13/source_target/in11.png}}\hfill
	\subfigure[]
	{\includegraphics[width=0.122\linewidth]{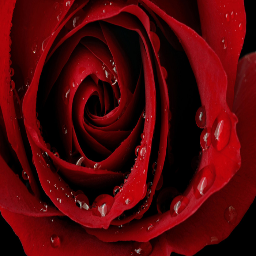}}\hfill
	\subfigure[]
	{\includegraphics[width=0.122\linewidth]{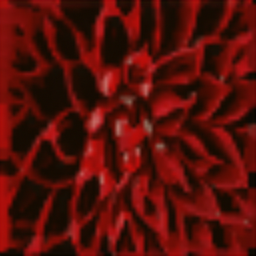}}\hfill
	\subfigure[]
	{\includegraphics[width=0.122\linewidth]{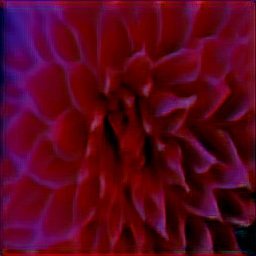}}\hfill
	\subfigure[]
	{\includegraphics[width=0.122\linewidth]{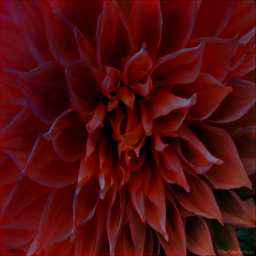}}\hfill
	\subfigure[]
	{\includegraphics[width=0.122\linewidth]{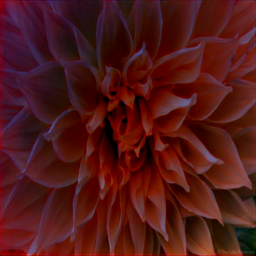}}\hfill
    \subfigure[]
	{\includegraphics[width=0.122\linewidth]{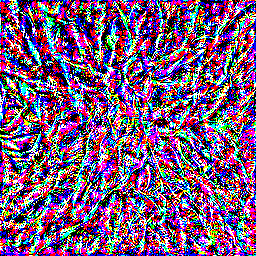}}\hfill
	\subfigure[]
	{\includegraphics[width=0.122\linewidth]{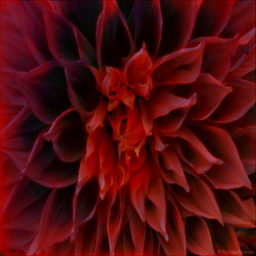}}\hfill\\
	\vspace{-20pt}

	\subfigure[]
	{\includegraphics[width=0.122\linewidth]{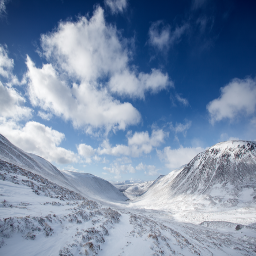}}\hfill
	\subfigure[]
	{\includegraphics[width=0.122\linewidth]{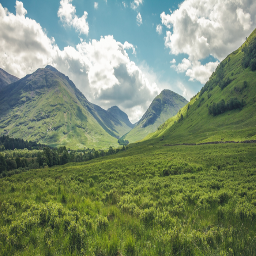}}\hfill
	\subfigure[]
	{\includegraphics[width=0.122\linewidth]{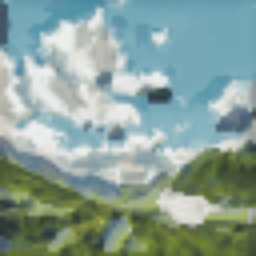}}\hfill
	\subfigure[]
	{\includegraphics[width=0.122\linewidth]{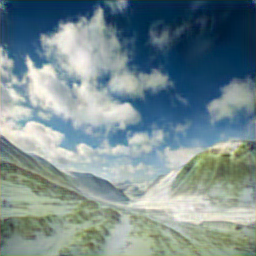}}\hfill
	\subfigure[]
	{\includegraphics[width=0.122\linewidth]{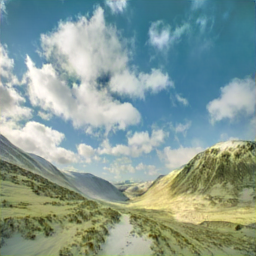}}\hfill
	\subfigure[]
	{\includegraphics[width=0.122\linewidth]{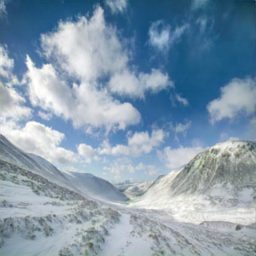}}\hfill
    \subfigure[]
	{\includegraphics[width=0.122\linewidth]{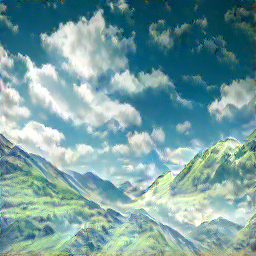}}\hfill
	\subfigure[]
	{\includegraphics[width=0.122\linewidth]{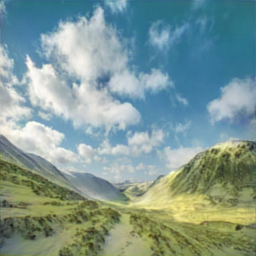}}\hfill\\
	\vspace{-20pt}

	\subfigure[]
	{\includegraphics[width=0.122\linewidth]{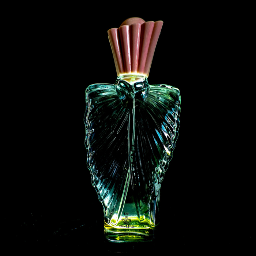}}\hfill
	\subfigure[]
	{\includegraphics[width=0.122\linewidth]{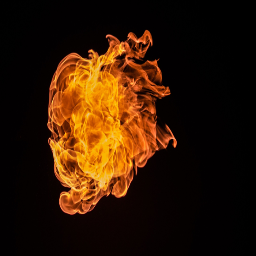}}\hfill
	\subfigure[]
	{\includegraphics[width=0.122\linewidth]{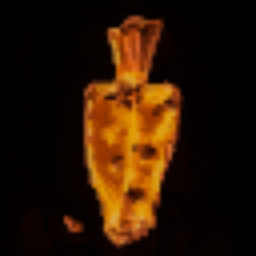}}\hfill
	\subfigure[]
	{\includegraphics[width=0.122\linewidth]{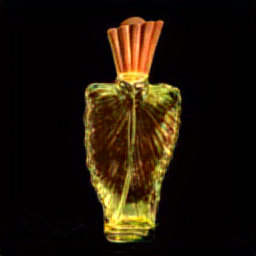}}\hfill
	\subfigure[]
	{\includegraphics[width=0.122\linewidth]{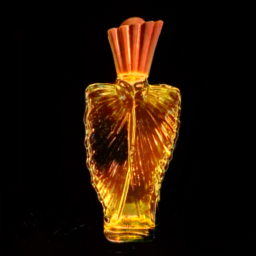}}\hfill
	\subfigure[]
	{\includegraphics[width=0.122\linewidth]{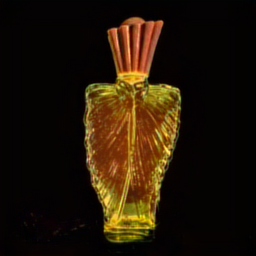}}\hfill
    \subfigure[]
	{\includegraphics[width=0.122\linewidth]{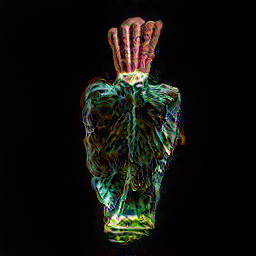}}\hfill
	\subfigure[]
	{\includegraphics[width=0.122\linewidth]{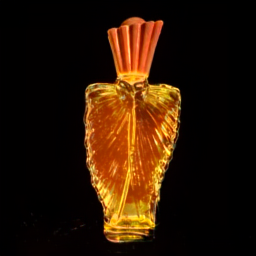}}\hfill\\
	\vspace{-20pt}
	
	\subfigure[]
	{\includegraphics[width=0.122\linewidth]{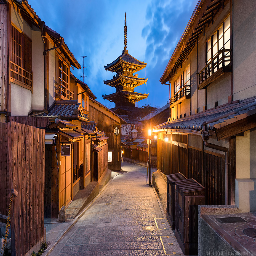}}\hfill
	\subfigure[]
	{\includegraphics[width=0.122\linewidth]{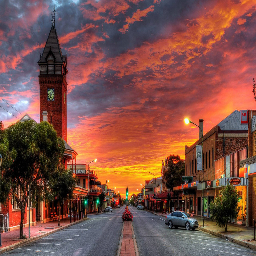}}\hfill
	\subfigure[]
	{\includegraphics[width=0.122\linewidth]{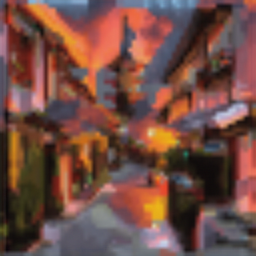}}\hfill
	\subfigure[]
	{\includegraphics[width=0.122\linewidth]{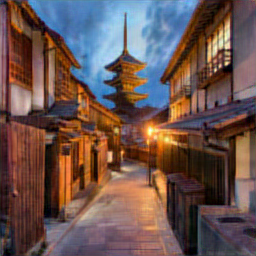}}\hfill
	\subfigure[]
	{\includegraphics[width=0.122\linewidth]{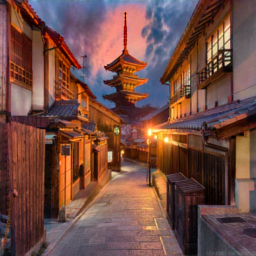}}\hfill
	\subfigure[]
	{\includegraphics[width=0.122\linewidth]{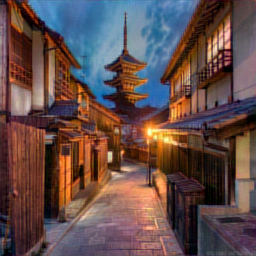}}\hfill
    \subfigure[]
	{\includegraphics[width=0.122\linewidth]{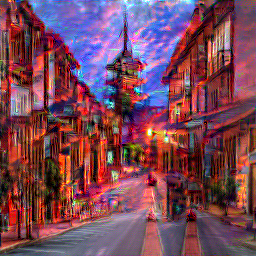}}\hfill
	\subfigure[]
	{\includegraphics[width=0.122\linewidth]{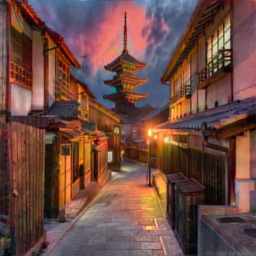}}\hfill\\
	\vspace{-20pt}

	\subfigure[]
	{\includegraphics[width=0.122\linewidth]{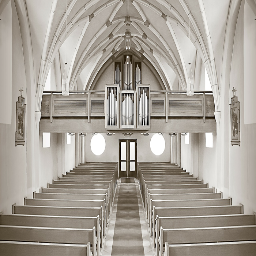}}\hfill
	\subfigure[]
	{\includegraphics[width=0.122\linewidth]{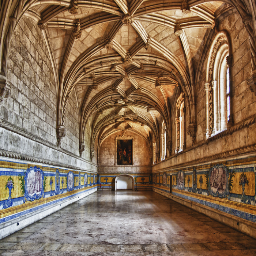}}\hfill
	\subfigure[]
	{\includegraphics[width=0.122\linewidth]{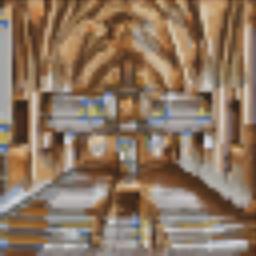}}\hfill
	\subfigure[]
	{\includegraphics[width=0.122\linewidth]{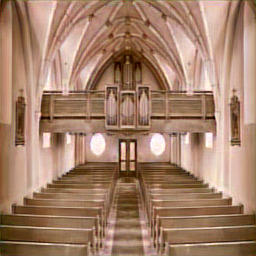}}\hfill
	\subfigure[]
	{\includegraphics[width=0.122\linewidth]{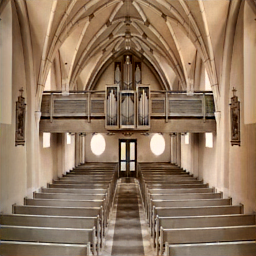}}\hfill
	\subfigure[]
	{\includegraphics[width=0.122\linewidth]{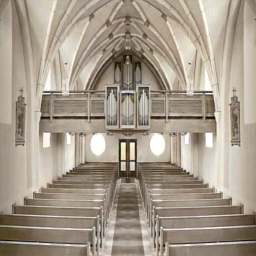}}\hfill
    \subfigure[]
	{\includegraphics[width=0.122\linewidth]{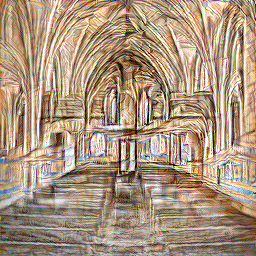}}\hfill
	\subfigure[]
	{\includegraphics[width=0.122\linewidth]{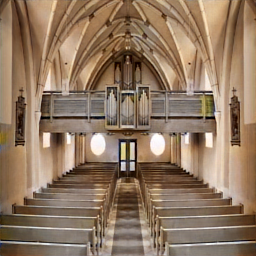}}\hfill\\
	\vspace{-20pt}

	\subfigure[Content]
	{\includegraphics[width=0.122\linewidth]{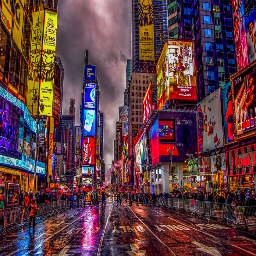}}\hfill
	\subfigure[Style]
	{\includegraphics[width=0.122\linewidth]{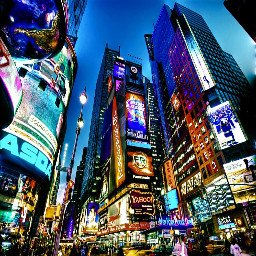}}\hfill
	\subfigure[w/o WF]
	{\includegraphics[width=0.122\linewidth]{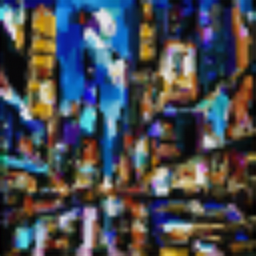}}\hfill
	\subfigure[w/o WI]
	{\includegraphics[width=0.122\linewidth]{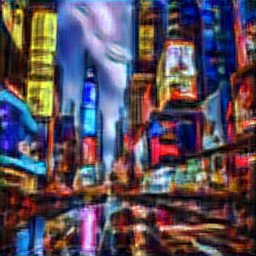}}\hfill
	\subfigure[w/o FMA]
	{\includegraphics[width=0.122\linewidth]{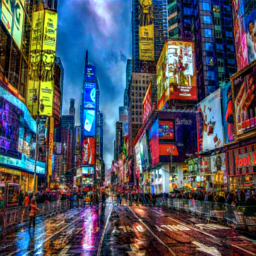}}\hfill
	\subfigure[w/o $\mathcal{L}_\mathrm{cyc}$]
	{\includegraphics[width=0.122\linewidth]{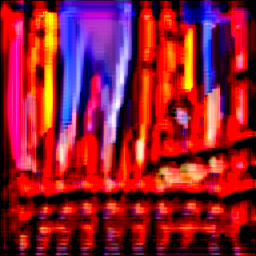}}\hfill
    \subfigure[w/o G. Module.]
	{\includegraphics[width=0.122\linewidth]{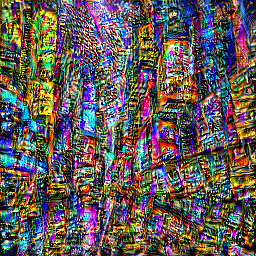}}\hfill
	\subfigure[Ours]
	{\includegraphics[width=0.122\linewidth]{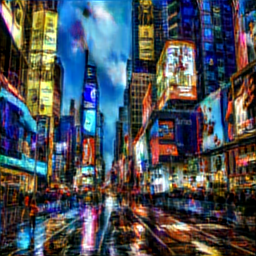}}\hfill\\
	\vspace{-10pt}
    \caption{\textbf{Additional examples on ablation study on priors, cycle consistency loss, generation module and feature moving average:} warped features (WF) and images (WI) help better converge. $\mathcal{L}_\mathrm{cyc}$ helps to optimization process to be stable and feature moving average (FMA) makes the results more stylized.} 
	\label{fig:supp_abl}\vspace{-10pt}
\end{figure*}

In the main paper, we have examined the impacts of i) warped feature (WF), ii) warped image (WI), iii) feature moving average (FMA), iv) cycle consistency regularization, and v) generation module (G. Module) in our framework.
We conduct the experiments with more examples from standard photorealistic style transfer~\cite{luan2017deep,an2019ultrafast}. The results are shown in \figref{fig:supp_abl}. 
We can observe the best stylization results of the final column. However, In the last two row, the better high frequency details are preserved in the w/o FMA than our final outputs. %\vspace{-5pt}
% ablation 결과에 대한 figure.

\subsection{Effects of Different Content-Style Weight.} 
\begin{figure*}[t]
\centering
	\renewcommand{\thesubfigure}{}
	\subfigure[]
	{\includegraphics[width=0.166\linewidth]{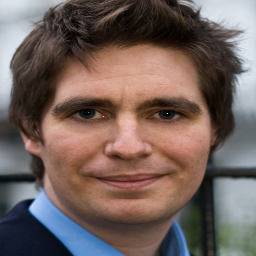}}\hfill
	\subfigure[]
	{\includegraphics[width=0.166\linewidth]{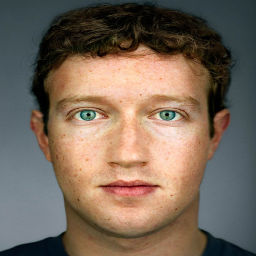}}\hfill
	\subfigure[]
	{\includegraphics[width=0.166\linewidth]{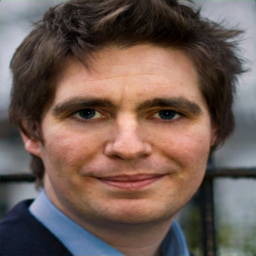}}\hfill
	\subfigure[]
	{\includegraphics[width=0.166\linewidth]{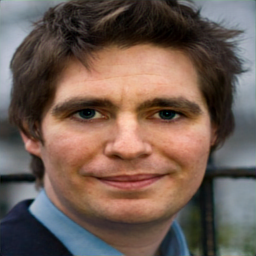}}\hfill
    \subfigure[]
	{\includegraphics[width=0.166\linewidth]{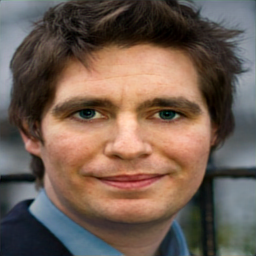}}\hfill
	\subfigure[]
	{\includegraphics[width=0.166\linewidth]{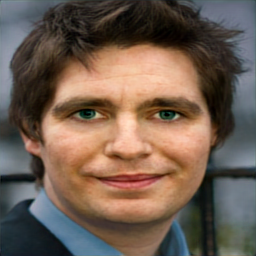}}\hfill\\
	\vspace{-20pt}	

	\subfigure[]
	{\includegraphics[width=0.166\linewidth]{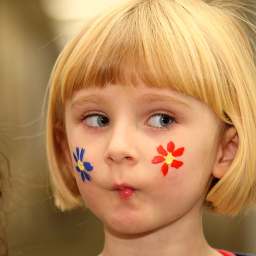}}\hfill
	\subfigure[]
	{\includegraphics[width=0.166\linewidth]{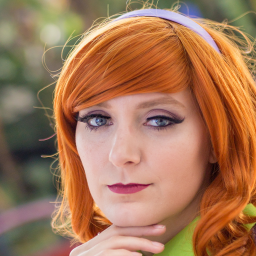}}\hfill
	\subfigure[]
	{\includegraphics[width=0.166\linewidth]{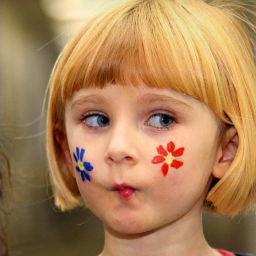}}\hfill
	\subfigure[]
	{\includegraphics[width=0.166\linewidth]{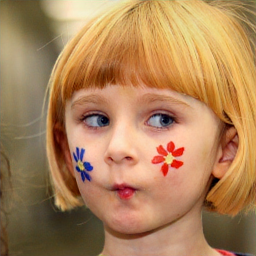}}\hfill
    \subfigure[]
	{\includegraphics[width=0.166\linewidth]{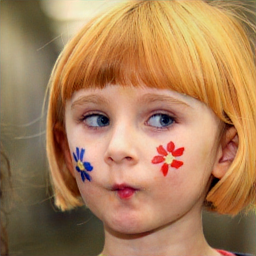}}\hfill
	\subfigure[]
	{\includegraphics[width=0.166\linewidth]{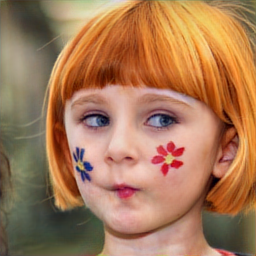}}\hfill\\
	\vspace{-20pt}	
	
	\subfigure[]
	{\includegraphics[width=0.166\linewidth]{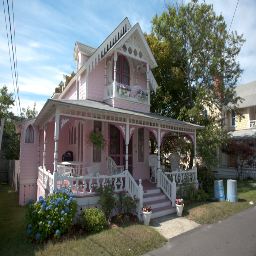}}\hfill
	\subfigure[]
	{\includegraphics[width=0.166\linewidth]{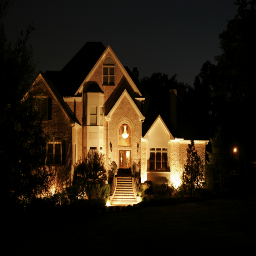}}\hfill
	\subfigure[]
	{\includegraphics[width=0.166\linewidth]{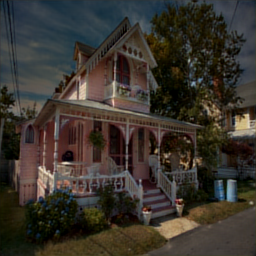}}\hfill
	\subfigure[]
	{\includegraphics[width=0.166\linewidth]{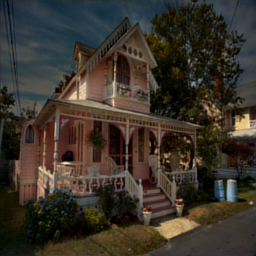}}\hfill
    \subfigure[]
	{\includegraphics[width=0.166\linewidth]{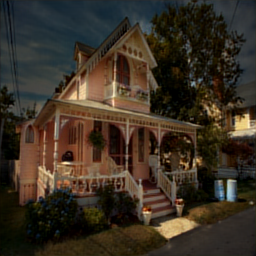}}\hfill
	\subfigure[]
	{\includegraphics[width=0.166\linewidth]{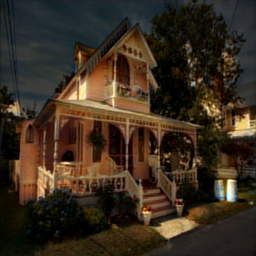}}\hfill\\
	\vspace{-20pt}	

	\subfigure[]
	{\includegraphics[width=0.166\linewidth]{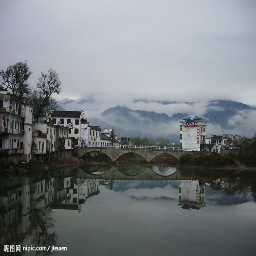}}\hfill
	\subfigure[]
	{\includegraphics[width=0.166\linewidth]{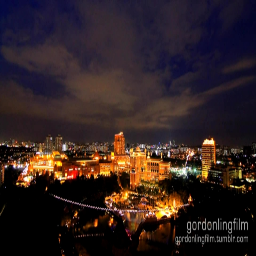}}\hfill
	\subfigure[]
	{\includegraphics[width=0.166\linewidth]{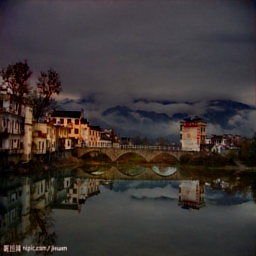}}\hfill
	\subfigure[]
	{\includegraphics[width=0.166\linewidth]{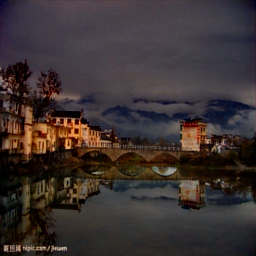}}\hfill
    \subfigure[]
	{\includegraphics[width=0.166\linewidth]{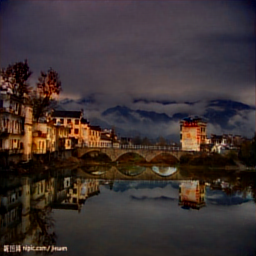}}\hfill
	\subfigure[]
	{\includegraphics[width=0.166\linewidth]{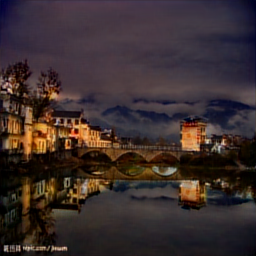}}\hfill\\
	\vspace{-20pt}

	\subfigure[]
	{\includegraphics[width=0.166\linewidth]{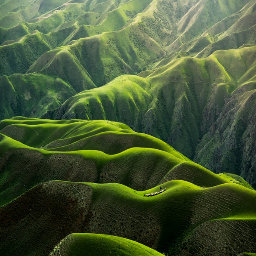}}\hfill
	\subfigure[]
	{\includegraphics[width=0.166\linewidth]{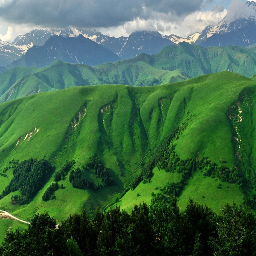}}\hfill
	\subfigure[]
	{\includegraphics[width=0.166\linewidth]{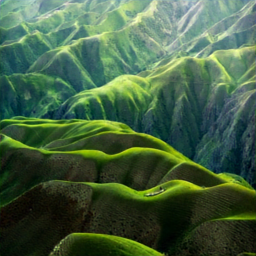}}\hfill
	\subfigure[]
	{\includegraphics[width=0.166\linewidth]{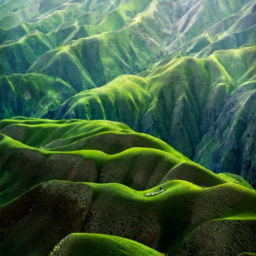}}\hfill
    \subfigure[]
	{\includegraphics[width=0.166\linewidth]{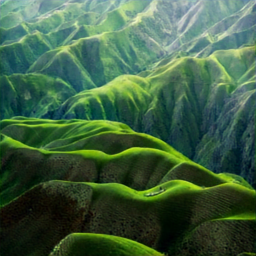}}\hfill
	\subfigure[]
	{\includegraphics[width=0.166\linewidth]{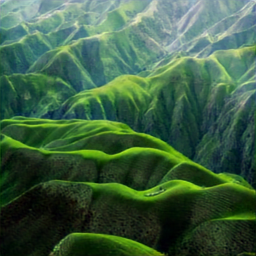}}\hfill\\
	\vspace{-20pt}	
	
	\subfigure[Content]
	{\includegraphics[width=0.166\linewidth]{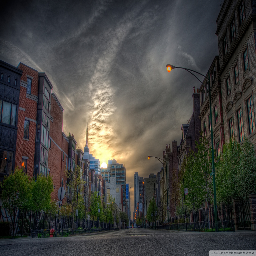}}\hfill
	\subfigure[Style]
	{\includegraphics[width=0.166\linewidth]{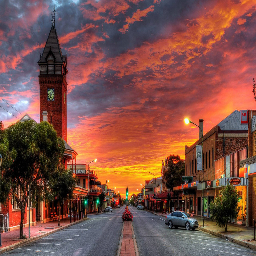}}\hfill
	\subfigure[$\lambda_{c}=1/5$]
	{\includegraphics[width=0.166\linewidth]{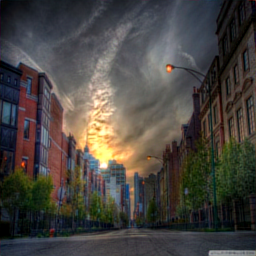}}\hfill
	\subfigure[$\lambda_{c}=2/5$]
	{\includegraphics[width=0.166\linewidth]{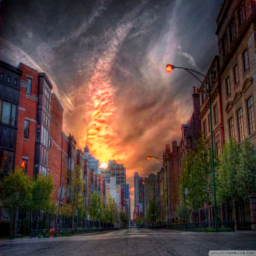}}\hfill
	\subfigure[$\lambda_{c}=3/5$]
	{\includegraphics[width=0.166\linewidth]{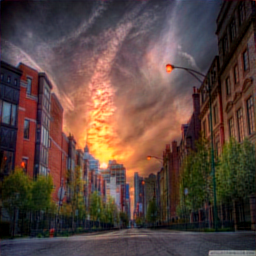}}\hfill
	\subfigure[$\lambda_{c}=4/5$]
	{\includegraphics[width=0.166\linewidth]{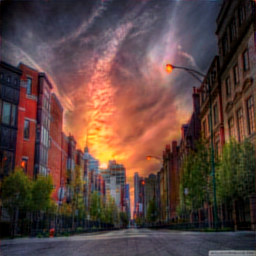}}\hfill\\
	\vspace{-10pt}
    \caption{\textbf{Examples of leveraging losses between content and style loss. The larger $\lambda_{c}$ gives more weight to the style term.}}
	\label{fig:supp_contentStyle}\vspace{-10pt}
\end{figure*}

In this section, we show the effect of changing weight of content and style loss. We set style weight as  $\lambda_\mathrm{c}$ and content weight as 1-$\lambda_\mathrm{c}$.
we vary the style weight as 1/5, 2/5, 3/5, 4/5 with default settings. Our results are shown in \figref{fig:supp_contentStyle}. Larger style weight tends to preserve more style details from style image, while larger content weight tends to produce results with more similar structure to content image.

\subsection{Effects of random seed.} 
\begin{figure*}[h]
	\begin{center}
	\renewcommand{\thesubfigure}{}
	\subfigure[]
	{\includegraphics[width=0.163\linewidth]{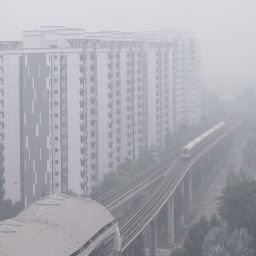}}\hfill %ph1 con
	\subfigure[]
	{\includegraphics[width=0.163\linewidth]{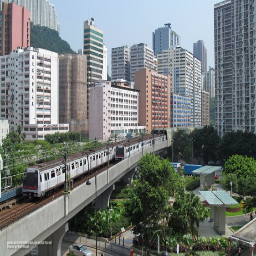}}\hfill
	\subfigure[]
	{\includegraphics[width=0.163\linewidth]{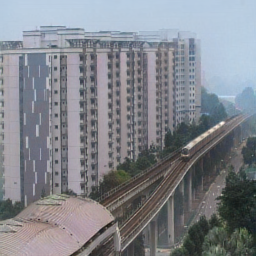}}\hfill %500
    \subfigure[]
	{\includegraphics[width=0.163\linewidth]{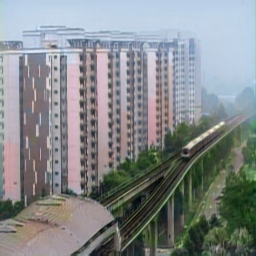}}\hfill %1500
	\subfigure[]	
	{\includegraphics[width=0.163\linewidth]{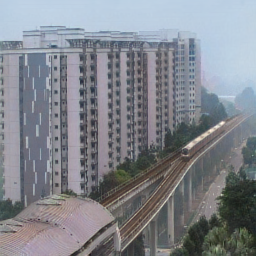}}\hfill %2000
	\subfigure[]	
	{\includegraphics[width=0.163\linewidth]{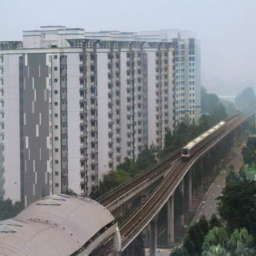}}\hfill\\  %1006
	\vspace{-20.5pt}
	\subfigure[]
	{\includegraphics[width=0.163\linewidth]{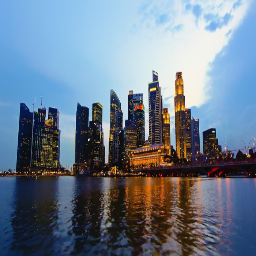}}\hfill
	\subfigure[]
	{\includegraphics[width=0.163\linewidth]{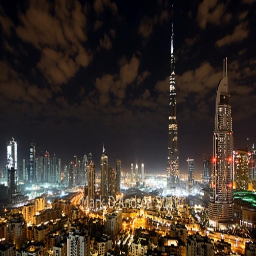}}\hfill
	\subfigure[]
	{\includegraphics[width=0.163\linewidth]{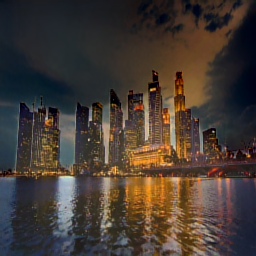}}\hfill %500
    \subfigure[]
	{\includegraphics[width=0.163\linewidth]{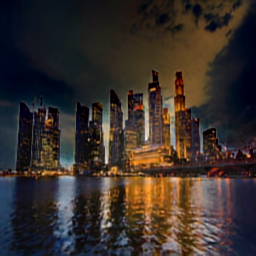}}\hfill %1500
	\subfigure[]	
	{\includegraphics[width=0.163\linewidth]{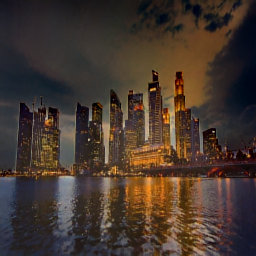}}\hfill %2000
	\subfigure[]	
	{\includegraphics[width=0.163\linewidth]{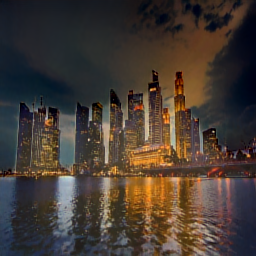}}\hfill\\  %1006
	\vspace{-20.5pt}
	\subfigure[]
	{\includegraphics[width=0.163\linewidth]{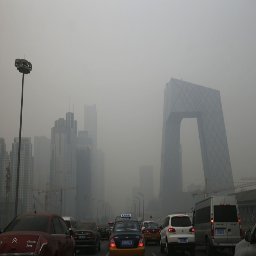}}\hfill
	\subfigure[]
	{\includegraphics[width=0.163\linewidth]{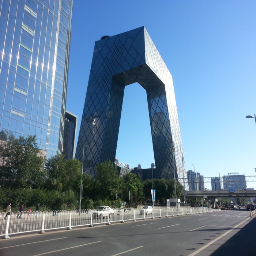}}\hfill
	\subfigure[]
	{\includegraphics[width=0.163\linewidth]{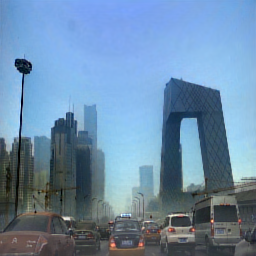}}\hfill %500
    \subfigure[]
	{\includegraphics[width=0.163\linewidth]{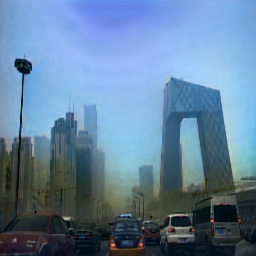}}\hfill %1500
	\subfigure[]	
	{\includegraphics[width=0.163\linewidth]{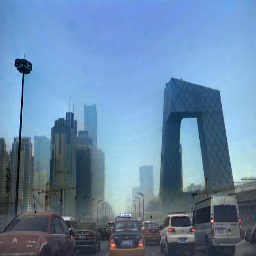}}\hfill %2000
	\subfigure[]	
	{\includegraphics[width=0.163\linewidth]{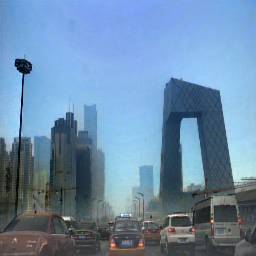}}\hfill\\  %1006
	\vspace{-20.5pt}
	\subfigure[]
	{\includegraphics[width=0.163\linewidth]{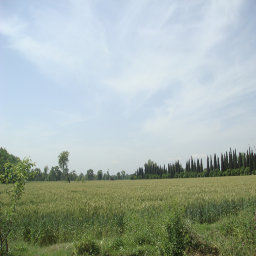}}\hfill
	\subfigure[]
	{\includegraphics[width=0.163\linewidth]{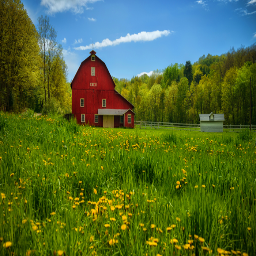}}\hfill
	\subfigure[]
	{\includegraphics[width=0.163\linewidth]{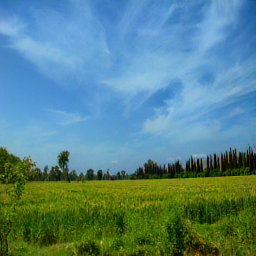}}\hfill %500
    \subfigure[]
	{\includegraphics[width=0.163\linewidth]{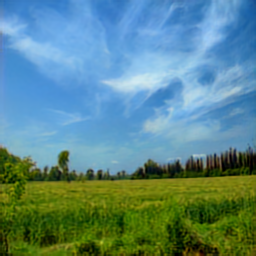}}\hfill %1500
	\subfigure[]	
	{\includegraphics[width=0.163\linewidth]{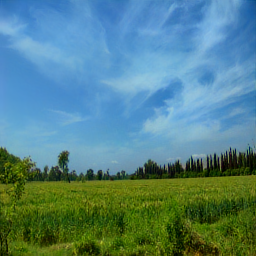}}\hfill %2000
	\subfigure[]	
	{\includegraphics[width=0.163\linewidth]{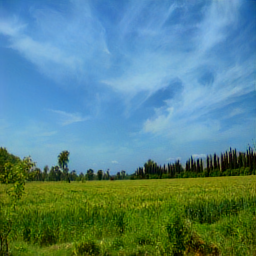}}\hfill\\  %1006
	\vspace{-20.5pt}
	\subfigure[]
	{\includegraphics[width=0.163\linewidth]{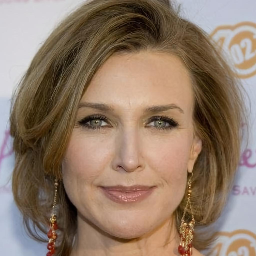}}\hfill
	\subfigure[]
	{\includegraphics[width=0.163\linewidth]{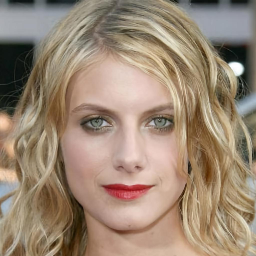}}\hfill
	\subfigure[]
	{\includegraphics[width=0.163\linewidth]{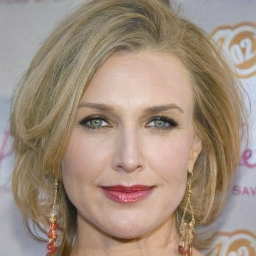}}\hfill %500
    \subfigure[]
	{\includegraphics[width=0.163\linewidth]{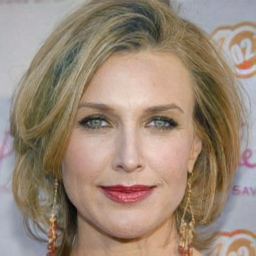}}\hfill %1500
	\subfigure[]	
	{\includegraphics[width=0.163\linewidth]{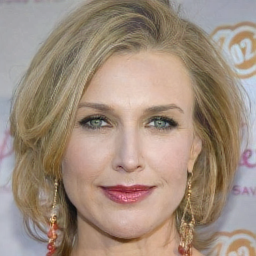}}\hfill %2000
	\subfigure[]	
	{\includegraphics[width=0.163\linewidth]{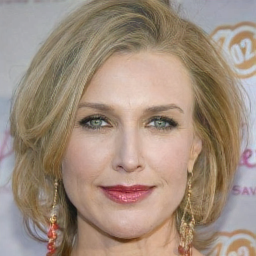}}\hfill\\  %1006
	\vspace{-20.5pt}

	\subfigure[]
	{\includegraphics[width=0.163\linewidth]{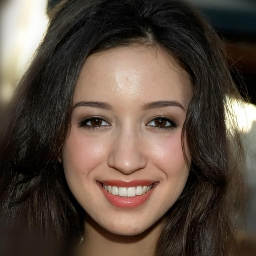}}\hfill
	\subfigure[]
	{\includegraphics[width=0.163\linewidth]{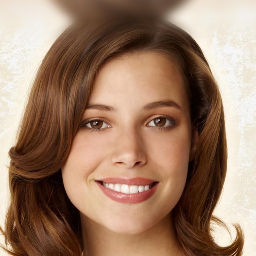}}\hfill
	\subfigure[]
	{\includegraphics[width=0.163\linewidth]{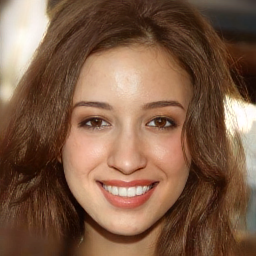}}\hfill %500
    \subfigure[]
	{\includegraphics[width=0.163\linewidth]{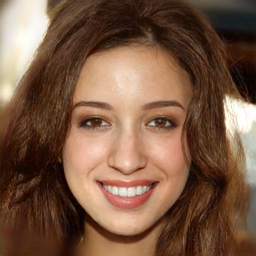}}\hfill %1500
	\subfigure[]	
	{\includegraphics[width=0.163\linewidth]{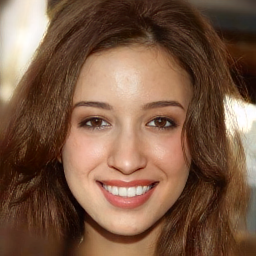}}\hfill %2000
	\subfigure[]	
	{\includegraphics[width=0.163\linewidth]{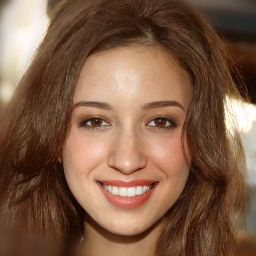}}\hfill\\  %1006
	\vspace{-20.5pt}

	\subfigure[]
	{\includegraphics[width=0.163\linewidth]{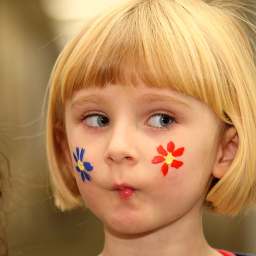}}\hfill
	\subfigure[]
	{\includegraphics[width=0.163\linewidth]{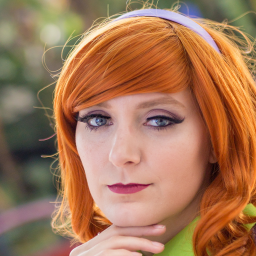}}\hfill
	{\includegraphics[width=0.163\linewidth]{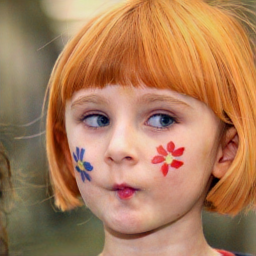}}\hfill %500
	\subfigure[]
	{\includegraphics[width=0.163\linewidth]{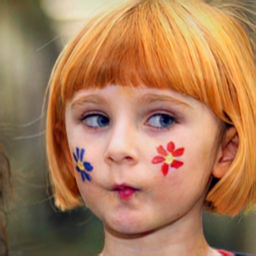}}\hfill %1500
	\subfigure[]	
	{\includegraphics[width=0.163\linewidth]{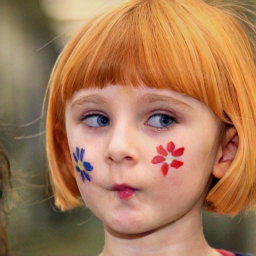}}\hfill %2000
	\subfigure[]
	{\includegraphics[width=0.163\linewidth]{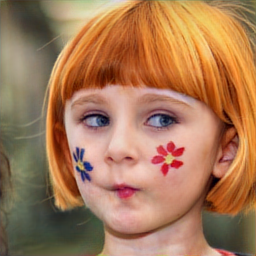}}\hfill\\  %1006
	\vspace{-20.5pt}	
	\subfigure[Content]
	{\includegraphics[width=0.163\linewidth]{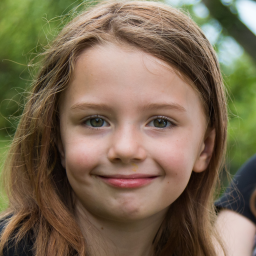}}\hfill
	\subfigure[Style]
	{\includegraphics[width=0.163\linewidth]{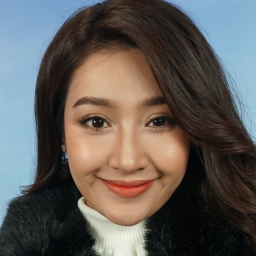}}\hfill
	\subfigure[500]
	{\includegraphics[width=0.163\linewidth]{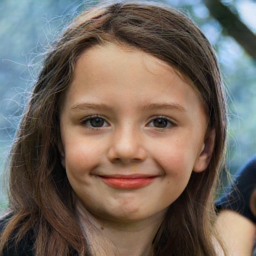}}\hfill %500
    \subfigure[1500]
	{\includegraphics[width=0.163\linewidth]{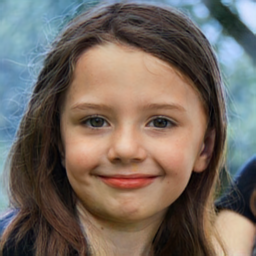}}\hfill %1500
	\subfigure[2000]
	{\includegraphics[width=0.163\linewidth]{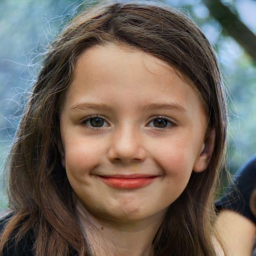}}\hfill %2000
	\subfigure[1006 (Ours)]		
	{\includegraphics[width=0.163\linewidth]{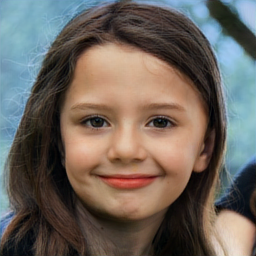}}\hfill\\  %1006	

	%\subfigure[\textbf{Ours-final}]
	%{\includegraphics[width=0.4\linewidth]{figure/ffhq_result/cft_ours/ours_4.png}}\hfill\\
    \vspace{-5pt}
	\caption{\textbf{Our results with various random seed.} }
	\label{fig:randomseed}\vspace{-10pt}
	    \end{center}
\end{figure*}

In this section, we show our results with respect to the random seed after running experiments multiple times. Our results are shown in \figref{fig:randomseed} with various kinds of random seed, which demonstrate that our model is robust to any random seed. %\vspace{-5pt}

\newpage

% content/style weight leveraging 결과에 대한 figure.

\end{document}